\documentclass[10pt]{article}

\usepackage{amsmath,amsfonts} 

\def\eqref#1{equation~\ref{#1}}

\def\1{\bm{1}}

\DeclareMathAlphabet{\mathsfit}{\encodingdefault}{\sfdefault}{m}{sl}
\SetMathAlphabet{\mathsfit}{bold}{\encodingdefault}{\sfdefault}{bx}{n}

\usepackage{arxiv_style}

\usepackage{graphicx}
\usepackage{tabularx}
\usepackage{booktabs}
\usepackage{multirow}
\usepackage{pifont}
\usepackage{enumitem}
\usepackage{wrapfig}
\title{TRIAGE: Direction-Aware Mismatch Stabilization of Native NVFP4 Reinforcement Learning}

\author{%
Zhen Li\textsuperscript{1,*},
Shuai Zhang\textsuperscript{2,*},
Yanggan Gu\textsuperscript{1},
Yiming Zhang\textsuperscript{1},
Yang Yu\textsuperscript{1},
Mingfa Feng\textsuperscript{1},\\
Congkai Xie\textsuperscript{1},
Shuang Yu\textsuperscript{2},
Junjie Lai\textsuperscript{2},
Hongxia Yang\textsuperscript{1}
}
\affiliation{\textsuperscript{1}InfiX.ai \quad \textsuperscript{2}NVIDIA}
\equalcontrib{\textsuperscript{*}These authors contributed equally to this work.}
\correspondence{\hypersetup{urlcolor=nv}\href{mailto:zhen.li@infix.io}{zhen.li@infix.io}, \hypersetup{urlcolor=nv}\href{mailto:shuazhang@nvidia.com}{shuazhang@nvidia.com}}
\code{\hypersetup{urlcolor=nv}\url{https://github.com/InfiXAI/TRIAGE}}
\logos{\raisebox{-0.5\height}{\includegraphics[height=8mm]{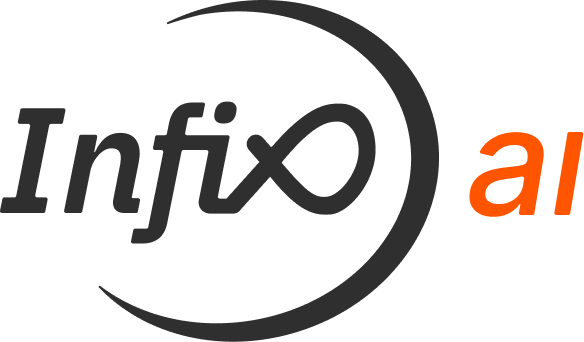}}}%
\date{September, 2026}
\abstract{%
Low-precision execution can substantially accelerate reinforcement learning
(RL) for large language models, but discrepancies between learner and sampler
execution can destabilize policy optimization.
In this paper, we characterize the interaction between mismatch and the
policy-gradient direction, distinguishing locally amplifying from
contracting update contributions that mismatch magnitude alone
cannot identify.
In native NVFP4 runs, we observe an early imbalance between the two
amplifying regions, favoring negative-advantage, negative-gap updates.
Their tail tokens become concentrated in a small fraction of response
segments before mismatch spreads globally.
Motivated by these findings, we introduce \textbf{TRIAGE}, a
direction-aware stabilization method that uses segment-level
diagnosis to selectively rebalance policy-gradient updates and
applies bounded repair to residual severe mismatch.
TRIAGE modifies the optimization objective while retaining native
NVFP4 weight-and-activation 4-bit (W4A4) forward execution on both
the sampler and learner.
Experiments on Qwen3-4B and Qwen3-30B-A3B show stable optimization
throughout the evaluated training horizon and achieve full precision level performance
across five mathematical reasoning benchmarks,
while native NVFP4 with TRIAGE provides up to 2.3$\times$ higher rollout throughput than BF16.
}

\begin{document}

\maketitle

\section{Introduction}
Reinforcement learning (RL) has become central to improving the reasoning capabilities of large language models (LLMs)~\citep{guo2025deepseekr1}. As RL expands toward long-horizon reasoning, coding, and agentic tasks, increasingly long generation trajectories make rollout a dominant wall-clock bottleneck of the training loop~\citep{qiu2026fp8rl,zhuge2026quads,xie2025infir}, making efficient rollout a primary systems challenge.

NVIDIA Blackwell GPUs provide native weight-and-activation 4-bit (W4A4) Tensor Core operations, offering up to $4\times$ the peak matrix-compute throughput of BF16 and making NVFP4 an attractive precision setting for long-sequence rollout~\citep{alvarez2025nvfp4,li2025quantization}. However, trajectories are sampled by one execution policy and optimized using probabilities recomputed by another. Differences between the rollout and training execution paths already induce learner--sampler probability discrepancies, and low-precision execution can further enlarge this mismatch~\citep{zhong2026diagnosing,gu2026qarl}. This creates a trade-off between hardware efficiency and stable policy optimization.

Existing approaches mitigate learner--sampler mismatch through two main
strategies. Quantization-aware training (QAT) attempts to align the learner
with the low-precision sampler~\citep{chen2025scaling}, but standard
 QAT retains high-precision arithmetic and therefore
does not preserve native W4A4 execution on the learner forward path.
Importance-based corrections such as truncated importance
sampling (TIS), clipping, and masking intervene according to the magnitude of
learner--sampler disagreement~\citep{qiu2026fp8rl,li2026qurl}.
More fundamentally, magnitude alone describes how far the learner and sampler
disagree, not whether the subsequent policy-gradient update will reduce or
further amplify that disagreement. This raises a more essential question:
\emph{what determines whether learner--sampler mismatch contracts or amplifies
under policy optimization?}
\begin{figure*}[tbp]
    \centering
    \includegraphics[width=\textwidth]{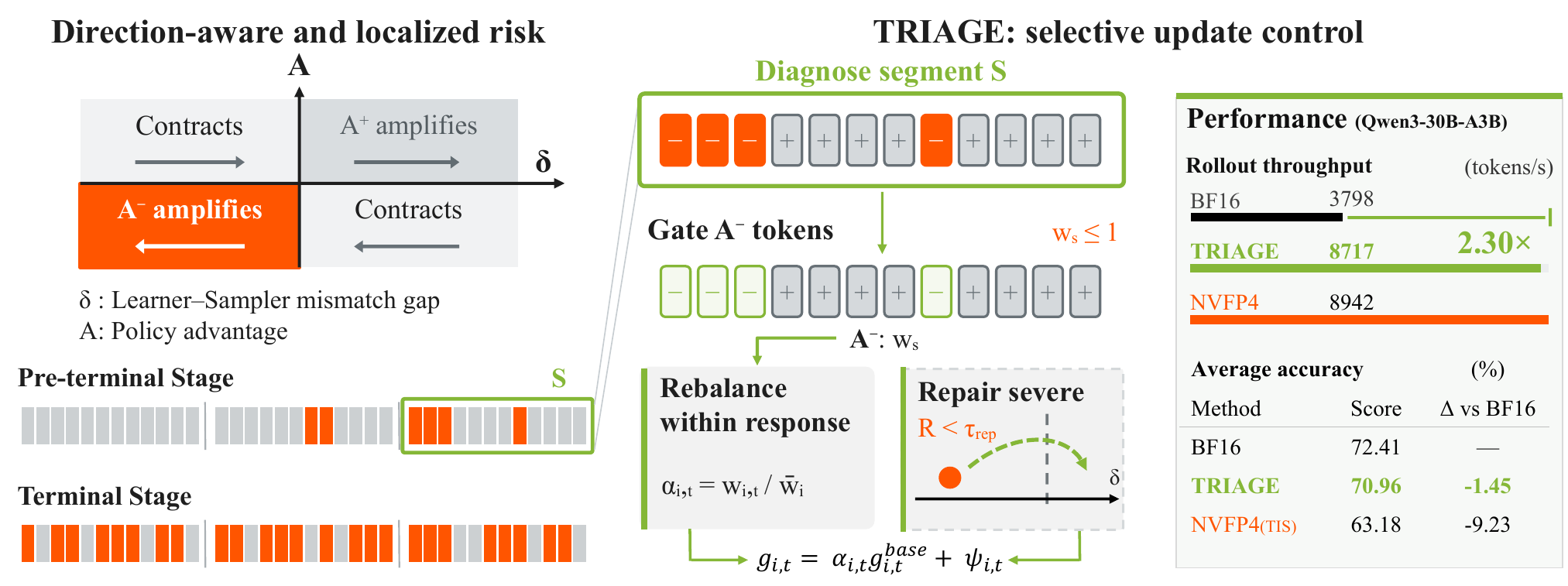}
    \caption{
    \textbf{Overview of TRIAGE for native NVFP4 RL.}
}
    \label{fig:main}
\end{figure*}

In this work, we show that learner--sampler mismatch must be
considered together with the direction of the policy update
(Figure~\ref{fig:main}, left).
If the learner already assigns a token a lower
probability than the sampler, further reducing that probability
can widen the mismatch rather than correct it.
In native NVFP4 runs, updates of this kind become increasingly
prominent relative to those that increase probabilities already
above the sampler's.
This imbalance appears before mismatch becomes broadly severe.
Tokens with larger discrepancies in this direction also become
concentrated within short portions of responses, where they can
be hidden by response-level averages.
These observations motivate a different stabilization principle:
\emph{identify and control updates that tend to amplify mismatch,
rather than intervening on mismatch magnitude alone}.

Motivated by these findings, we propose \textbf{TRIAGE}, a direction-aware stabilization framework for native low-precision RL. As illustrated in Figure~\ref{fig:main}, TRIAGE uses segment-level statistics to identify locally persistent risk, selectively attenuates only mismatch-amplifying \(\mathcal A^-\) (Advantage $<$ 0, mismatch gap $<$ 0) tokens within risky segments, rebalances update mass, and applies a bounded repair term to residual severe mismatch. By acting on the optimization objective rather than replacing low-precision execution with fake-quantized computation, TRIAGE retains native NVFP4 W4A4 forward execution on both the sampler and learner.
Across Qwen3-4B and training-sensitive Qwen3-30B-A3B MoE models, NVFP4 with TRIAGE achieves
up to 2.3$\times$ the rollout throughput of BF16, while maintaining stable optimization
throughout the training horizon and recovering comparable
performance on mathematical reasoning benchmarks.
Our contributions are summarized as follows:
\begin{itemize}[
    leftmargin=*,
    itemsep=5pt,
    topsep=2pt,
    parsep=0pt,
    partopsep=0pt,
    ]
   \item[\ding{182}]\textbf{A directional characterization of learner--sampler mismatch.}
   We characterize mismatch risk through its interaction with the policy-gradient direction and identify a directional, locally concentrated precursor to native NVFP4 instability.

    \item[\ding{183}] \textbf{TRIAGE, a selective stabilization mechanism for low-precision RL.}
    TRIAGE combines segment-level risk diagnosis, direction-selective token gating, response-level update rebalancing, and targeted repair to suppress self-amplifying mismatch while preserving contracting updates.

    \item[\ding{184}] \textbf{An end-to-end native NVFP4 RL recipe at scale.}
    Across experiments totaling more than 34,500 GPU-hours on B300 GPUs, we validate native W4A4 execution on both sampler and learner forward paths across dense and MoE models, achieving up to 1.3$\times$ end-to-end RL speedup while maintaining the BF16-level performance.

\end{itemize}
\section{Preliminaries}\label{sec:preliminaries}
\subsection{Group-Relative Policy Optimization}
Group-Relative Policy Optimization (GRPO)~\citep{shao2024deepseekmath} is widely used in RL for reasoning models. Given a prompt $x$, GRPO samples a group of $G$ responses
$\{y_i\}_{i=1}^{G}$ from an old policy $\pi_{\mathrm{old}}$ and evaluates their rewards $\{r_i\}_{i=1}^{G}$. 
For outcome-based RL, the group-relative advantage is typically
$A_i=(r_i-\mu_r)/(\sigma_r+\epsilon)$, where $\mu_r$ and $\sigma_r$ are the
mean and standard deviation of group rewards, and is assigned to all valid
tokens of response $i$.
Let 
$\rho_{i,t}=\pi_{\theta}(y_{i,t}\mid x,y_{i,<t})/
\pi_{\mathrm{old}}(y_{i,t}\mid x,y_{i,<t})$
donote the token-level importance ratio.
Ignoring optional regularization terms, the clipped policy objective can be written as
\begin{equation}
    \mathcal{L}_{\mathrm{GRPO}}
    =
    -\frac{1}{G}\sum_{i=1}^{G}\frac{1}{|y_i|}
    \sum_{t=1}^{|y_i|}
    \min\!\left(
        \rho_{i,t} A_i,\,
        \operatorname{clip}(\rho_{i,t},1-\varepsilon,1+\varepsilon)A_i
    \right),
\end{equation}
Locally, a positive advantage increases the probability of sampled tokens, whereas a negative advantage decreases it; this directional property will be central to our analysis in Section~\ref{sec:understanding}~\citep{qiu2026fp8rl,zhuge2026quads}.

\subsection{Decoupled Rollout and Training}
Modern LLM RL systems typically decouple rollout generation from gradient computation: a high-throughput inference engine such as SGLang or vLLM~\citep{zheng2024sglang,kwon2023efficient} acts as the \emph{sampler}, while a training framework such as Megatron-LM acts as the \emph{learner}~\citep{qiu2026fp8rl,shoeybi2019megatron}. Let
$\pi_{\mathrm{s}}$ and $\pi_{\mathrm{l}}$ denote the sampler and learner policies, respectively. For a sampled token $y_t$ under the same prefix $y_{<t}$, we define their log-probability gap as
\begin{equation}
    \delta_t
    =
    \log \pi_{\mathrm{l}}(y_t\mid x,y_{<t})
    -
    \log \pi_{\mathrm{s}}(y_t\mid x,y_{<t}).
    \label{eq:logp_gap}
\end{equation}
Ideally, synchronized policies should yield $\delta_t=0$. In practice, the two execution paths can disagree even when they nominally represent the same model~\citep{zhong2026diagnosing}.

This gap admits a useful interpretation in the language of off-policy
correction. Because sampled tokens are drawn from the sampler,
$y_{i,t}\sim\pi_{\mathrm{s}}(\cdot\mid x,y_{i,<t})$, the importance weight
that exact policy evaluation would require is not
$\rho_{i,t}=\pi_{\theta}/\pi_{\mathrm{old}}$ but
$\pi_{\theta}/\pi_{\mathrm{s}}$, which factors as
\begin{equation}
    \frac{\pi_{\theta}(y_{i,t}\mid x,y_{i,<t})}
         {\pi_{\mathrm{s}}(y_{i,t}\mid x,y_{i,<t})}
    =
    \underbrace{\frac{\pi_{\theta}(y_{i,t}\mid x,y_{i,<t})}
         {\pi_{\mathrm{l}}(y_{i,t}\mid x,y_{i,<t})}}_{\text{optimization drift}}
    \cdot
    \underbrace{\exp\!\big(\delta_{i,t}\big)}_{\text{learner--sampler ratio}}.
    \label{eq:ratio_factorization}
\end{equation}
Under synchronous weight transfer with one update per rollout, the first
factor is one at the start of each iteration, and the entire off-policy error
of the update is carried by $\exp(\delta_{i,t})$. Truncated importance
sampling, clipping, and masking~\citep{qiu2026fp8rl,li2026qurl} all
intervene on exactly this factor. Crucially, $\exp(\delta_{i,t})$ is a
\emph{magnitude-only} statistic: it records how far the two policies
disagree, but not how the ensuing gradient will change that disagreement.
This gap between what existing corrections measure and what optimization
stability requires motivates our directional analysis in
Section~\ref{sec:understanding}.

\subsection{Low-Precision Rollout with NVFP4}
Reduced-precision rollout lowers both matrix-computation cost and model-memory traffic, making it particularly attractive for long-sequence RL. However, quantization introduces a forward perturbation that is qualitatively different from ordinary policy staleness~\citep{li2025quantization}. Even under an identical prefix, quantized weights and activations perturb hidden states and output logits, producing a local probability discrepancy between the sampler and learner. During decoding, such local perturbations can additionally alter token sampling and hence the subsequent prefix, allowing quantization error to accumulate into trajectory-level divergence~\citep{zhuge2026quads,gu2026qarl}.

NVFP4 is a Blackwell-supported format combining E2M1 FP4 values,
an E4M3 FP8 microscale per 16 values, and an FP32 second-level
scale~\citep{nvidia2025nvfp4}. For an element $x_j$ in a 16-element block $b$, its quantized representation can be abstracted as
\begin{equation}
    \hat{x}_j
    =
    S\,s_b\,
    Q_{\mathrm{E2M1}}
    \left(
        \frac{x_j}{S\,s_b}
    \right),
    \qquad j\in b,
\end{equation}
where $s_b$ is an E4M3 FP8 microscale shared by 16 consecutive values, $S$ is an FP32 second-level scale shared across the tensor, and $Q_{\mathrm{E2M1}}(\cdot)$ denotes rounding to the FP4 E2M1 grid. The fine-grained E4M3 microscale improves local range utilization, while the tensor-level FP32 scale extends the range representable by the collection of FP8 block scales.

\section{Understanding Learner--Sampler Policy Mismatch}
\label{sec:understanding}
We now study how the learner--sampler mismatch introduced in
Section~\ref{sec:preliminaries} evolves under policy optimization. 
Our analysis reveals a staged failure process.
Section~\ref{sec:mismatch_dynamics} derives which mismatches are
locally amplified by the policy-gradient update.
Section~\ref{sec:directional_asymmetry} then identifies when this
directional risk becomes visible along training.
Finally, Section~\ref{sec:localized_mismatch} shows where the
risk accumulates within long responses and why response-level
averages fail to expose it.

\subsection{Directional Dynamics of Policy Mismatch}
\label{sec:mismatch_dynamics}
\paragraph{Which mismatches are self-amplifying under an RL policy update?} 
Let $z_{i,t}(\theta)=\log\pi_\theta(y_{i,t}\mid x,y_{i,<t})$ denote the
learner log-probability of rollout token $(i,t)$, $\delta_{i,t}$ is log-probability gap from \eqref{eq:logp_gap}, and $g_{i,t}$ the signed coefficient applied to its policy-gradient term. A generic learner
update with step size $\eta > 0$ is
\begin{equation}
    \theta^{+}
    =
    \theta
    +
    \eta
    \sum_{i,t}
    g_{i,t}\nabla_\theta z_{i,t}(\theta),
    \label{eq:generic_policy_update}
\end{equation}
where for vanilla GRPO at the start of a single-update iteration
$g_{i,t}=A_i$; importance correction and stabilization mechanisms modify
this coefficient. To track whether the mismatch grows or contracts,
define the mismatch energy $D=\frac{1}{2}\sum_{i,t}\delta_{i,t}^{2}$.
Holding the rollout policy fixed during one update, a first-order expansion gives
\begin{equation}
    \Delta D
    =
    \eta\,\boldsymbol{\delta}^{\top}K\mathbf{g}
    +
    \mathcal{O}(\eta^2)
    =
    \eta
    \sum_{i,t}
    \kappa_{i,t}\delta_{i,t}g_{i,t}
    +
    \eta R_{\mathrm{cross}}
    +
    \mathcal{O}(\eta^2),
    \label{eq:mismatch_dynamics}
\end{equation}
where $K$ is the Gram matrix of token log-probability gradients,
$\kappa_{i,t}=\|\nabla_\theta z_{i,t}\|_2^2$, and $R_{\mathrm{cross}}$
collects cross-token interactions. Thus $\Delta D>0$ means the current
update batch inflates the total mismatch energy. 
Appendix~\ref{app:mismatch_derivation} derives this expansion for a
general preconditioned update with explicit optimizer residuals.
Equation \ref{eq:mismatch_dynamics} is the identity-preconditioner special
case. 

Writing the amplifying and contracting diagonal masses as
\begin{equation}
    B_{\mathrm{amp}}=\sum_{i,t}\kappa_{i,t}[\delta_{i,t}g_{i,t}]_{+},  \qquad
    B_{\mathrm{con}}=\sum_{i,t}\kappa_{i,t}[-\delta_{i,t}g_{i,t}]_{+}
\end{equation}
where $[x]_{+}=\max(x,0)$, it becomes
\begin{equation}
    \Delta D
    =
    \eta
    \left(
        B_{\mathrm{amp}}
        -
        B_{\mathrm{con}}
        +
        R_{\mathrm{cross}}
    \right)
    +
    \mathcal{O}(\eta^2).
    \label{eq:amp_con_balance}
\end{equation}
Thus, within the diagonal current-gradient contribution,
$\delta_{i,t}g_{i,t}>0$ amplifies the existing mismatch, whereas
$\delta_{i,t}g_{i,t}<0$ contracts it. In particular, for vanilla GRPO,
the criterion reduces to the sign interaction $A_i\delta_{i,t}$.
The two mismatch-amplifying regions are
\begin{equation}
\mathcal{A}^{-}
=
\{(i,t): A_i<0,\ \delta_{i,t}<0\},
\qquad
\mathcal{A}^{+}
=
\{(i,t): A_i>0,\ \delta_{i,t}>0\}.
\label{eq:amplifying_regions}
\end{equation}
The remaining two sign combinations are locally
mismatch-contracting. Figure~\ref{fig:main}, left summarizes
this directional geometry.

This distinction explains why mismatch magnitude
alone is insufficient: \textbf{a large mismatch may already be self-correcting,
while a moderate one grows systematically when the gradient acts along it.}
Cross-token coupling, optimizer-history terms, and higher-order effects
can still alter the realized motion of an individual token, so we use
$\delta_{i,t}g_{i,t}$ as a local directional risk indicator rather than
an unconditional predictor of the realized sign of
$\delta_{i,t}\Delta\delta_{i,t}$. Appendix~\ref{app:directional_validity} 
audits the scope of this interpretation. Our main use of the decomposition is
therefore population-level: it predicts that instability may first
appear as an imbalance between the two amplifying regions, even before
the marginal gap distribution becomes broadly severe.

\subsection{Directional Asymmetry Precedes Negative-Tail Explosion}
\label{sec:directional_asymmetry}
\paragraph{When does the imbalance become visible?}
We characterize native-NVFP4 GRPO trajectories of
Qwen3-4B-Base and Qwen3-30B-A3B-Base using four 25-update trailing
windows placed at 25\%, 50\%, 75\%, and 100\% of each run's
characterization horizon. 
We refer to these windows descriptively as
\emph{healthy}, \emph{drift}, \emph{pre-terminal}, and
\emph{terminal}, respectively.

To measure the distribution of tokens between the two
amplifying regions in \eqref{eq:amplifying_regions}, define the
advantage-weighted token occupancy
\begin{equation}
w_A(\mathcal{R})
=
\frac{
\sum_{(i,t)\in\mathcal{R}} |A_i|
}{
\sum_{i,t}|A_i|
},
\quad
\rho_{\mathrm{asym}}
=
\frac{
w_A(\mathcal{A}^{-})
}{
w_A(\mathcal{A}^{+})
}.
\label{eq:asymmetry_ratio}
\end{equation}
\begin{wrapfigure}[14]{r}{0.4\columnwidth}
    \centering
    \includegraphics[width=\linewidth]{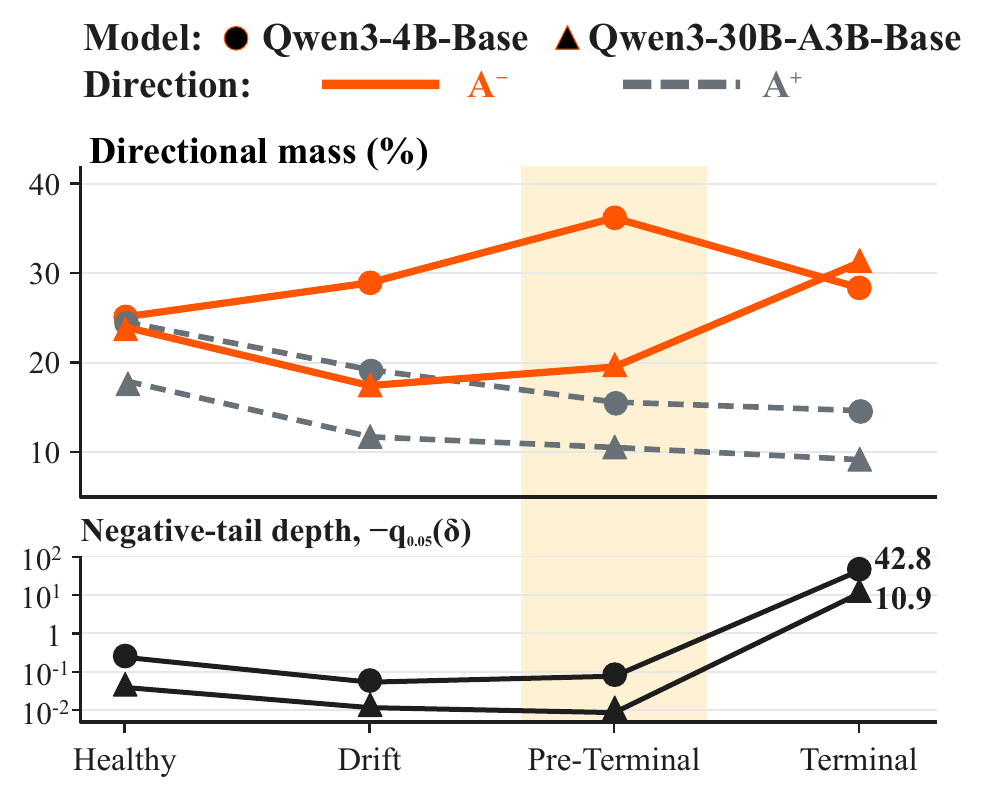}
    \caption{Directional asymmetry precedes terminal negative-tail expansion}
    \label{fig:mismatch_analysis}
\end{wrapfigure}
Figure~\ref{fig:mismatch_analysis} reveals a separation between
directional drift and magnitude collapse. Across the
drift-to-pre-terminal interval, the marginal gap distribution still
appears largely benign: the fraction of tokens satisfying
$|\delta|<0.05$ remains at least 83.0\% on 4B and 91.9\% on 30B.
Nevertheless, $\rho_{\mathrm{asym}}$ rises from 1.02 to 2.32 by pre-terminal on 4B, 
and from 1.34 to 1.85 on 30B.
The two theoretically symmetric amplifying regions therefore become
empirically imbalanced well before a broad magnitude increase is
visible. 

In the terminal stage, the lower tail expands abruptly.
Within the respective mismatch estimands, the 5\% gap quantile 
$q_{0.05}(\delta)$ reaches $-42.75$ on 4B and
$-10.94$ on 30B. \textbf{Directional asymmetry therefore identifies an
earlier warning interval, whereas the negative-tail explosion
characterizes the terminal regime.}

\subsection{Localized Accumulation before Global Failure}
\label{sec:localized_mismatch}
\paragraph{Where does the amplifying mismatch accumulate?}
For each negative-advantage response, 
we partition it into non-overlapping segments $S$ of
$W=64$ tokens. 
We call tokens satisfying $A_i<0$ and
$\delta_{i,t}<\tau_{\mathrm{ana}}$ \emph{amplifying-tail tokens},
where the model-specific threshold
$\tau_{\mathrm{ana}}^{(m)}$ is fixed as the fifth percentile of
$\delta$ in the corresponding 
\begin{wrapfigure}[20]{r}{0.43\columnwidth}
    \centering
    \setlength{\abovecaptionskip}{0pt}
    \includegraphics[width=\linewidth]{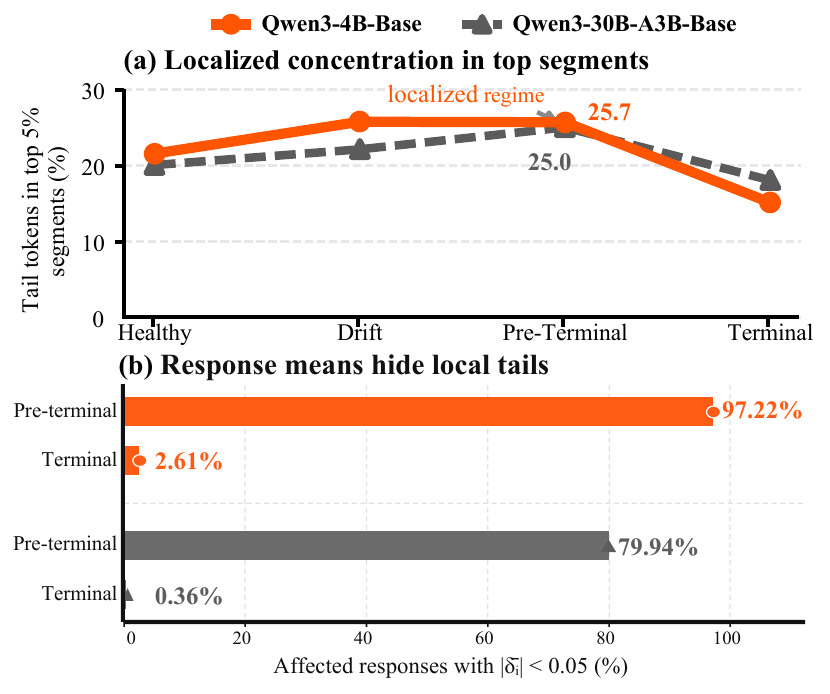}
    \caption{\textbf{Localized mismatch is hidden by response means before terminal.}
    (a) Share of amplifying-tail tokens contained in the top 5\% of
    segments ranked by $f_S^{\mathrm{amp}}$.
    (b) Fraction of negative-advantage responses containing at least one tail-heavy segment, the fraction with $|\bar\delta_i|<0.05$}
    \label{fig:server_tail}
    \vspace{2pt}
\end{wrapfigure}
healthy window.
Their local concentration is
measured by
\begin{equation}
    f_S^{\mathrm{amp}}(\tau_{\mathrm{ana}})
    =
    \frac{1}{|S|}
    \sum_{t\in S}
    \mathbf{1}[A_i<0,\ \delta_{i,t}<\tau_{\mathrm{ana}}],
    \label{eq:segment_amp_tail}
\end{equation}
we call a segment \emph{tail-heavy} when
$f_S^{\mathrm{amp}}>0.1$.

From the healthy to the pre-terminal window, their global frequency
decreases from $4.8$--$4.9\%$ to $2.8$--$3.6\%$ across the two
models. Nevertheless, the top $5\%$ of segments contain approximately
one quarter of all amplifying-tail tokens, about five times the mass
implied by uniform placement, while adjacent tail-heavy persistence
increases from approximately $2.6\times$ to
$3.8$--$4.0\times$ the chance rate.
The precursor is therefore not an expanding global tail, but a
reorganization of a small tail population into a few persistent local
regions.

Response averages obscure this structure: at pre-terminal,
$97.2\%$ (4B) and $79.9\%$ (30B) of responses containing
tail-heavy segments still satisfy $|\bar\delta_i|<0.05$,
where $\bar\delta_i=n_i^{-1}\sum_{t\in\mathcal T_i}\delta_{i,t}$.
By terminal, roughly half of the eligible segments are tail-heavy
and top-segment concentration declines.

These observations motivate segment-level diagnosis with
intervention restricted to $\mathcal A^-$ tokens, rather than
suppressing every token in an affected response or segment.

\section{Method}
\label{sec:method}
Section~\ref{sec:understanding} shows that the risk of learner--sampler
mismatch depends on its interaction with the policy-update direction:
updates with $\delta_{i,t}g_{i,t}>0$ amplify the existing gap, whereas
updates with $\delta_{i,t}g_{i,t}<0$ contract it. 
TRIAGE acts on this distinction through two complementary modifications of the effective update coefficient. Let $g^{\mathrm{base}}_{i,t}$ denote the per-token coefficient of the underlying importance-corrected objective, including its reduction factors. We then write
\begin{equation}
    g_{i,t} = \alpha_{i,t}\, g^{\mathrm{base}}_{i,t} + \psi_{i,t},
    \label{eq:TRIAGE_update}
\end{equation}
The two terms partition by advantage sign: $\alpha_{i,t}$
reallocates policy-gradient mass within negative-advantage responses
(Section~\ref{sec:TRIAGE_gate}), while $\psi_{i,t}$ adds bounded
corrective pressure on positive-advantage responses
(Section~\ref{sec:TRIAGE_repair}). 
\subsection{Segment Diagnosis with Direction-Selective Gating}
\label{sec:TRIAGE_gate}
Following Section~3.3, TRIAGE therefore uses short contiguous segments as an intermediate
diagnostic scale, a granularity that has also been found useful for
avoiding the extremes of token- and response-level importance
estimation~\citep{yang2026sspo}.

For a diagnostic segment $S$ of response $i$, we compute
\begin{equation}
    \bar{\delta}_S
    =
    \frac{\sum_{t\in S}m_{i,t}\delta_{i,t}}
         {\sum_{t\in S}m_{i,t}},
    \qquad
    f^{\mathrm{tail}}_S
    =
    \frac{\sum_{t\in S}
        m_{i,t}\mathbf{1}
        [\delta_{i,t}<\tau_{\mathrm{tail}}]}
         {\sum_{t\in S}m_{i,t}},
    \label{eq:segment_statistics}
\end{equation}
which respectively capture persistent displacement and sparse deep-tail
mismatch. We map these statistics to a segment gate weight
\begin{equation}
    w_S
    =
    \mathcal{G}
    \left(
        \bar{\delta}_S,
        f^{\mathrm{tail}}_S
    \right)
    \in [w_{\min},1],
    \label{eq:segment_risk_map}
\end{equation}
where a larger negative displacement or a heavier negative tail yields
stronger attenuation. The exact operating rule, including eligibility
conditions and the fixed operating thresholds, is deferred to
Appendix~\ref{app:gate_details}.

Crucially, a risky segment does not imply that all of its tokens should be
attenuated. For a negative-advantage response, tokens with
$A_i<0,\ \delta_{i,t}<0$ contribute to mismatch amplification, whereas
tokens with $A_i<0,\ \delta_{i,t}>0$ naturally contract the discrepancy.
TRIAGE therefore applies
\begin{equation}
    w_{i,t}
    =
    \begin{cases}
        w_S,
        & A_i<0,\ \delta_{i,t}<0,\ t\in S,\\[2pt]
        1,
        & \text{otherwise}.
    \end{cases}
    \label{eq:direction_selective_gate}
\end{equation}
Thus, segment statistics determine \emph{whether} local intervention is
warranted, while the directional criterion from
Section~\ref{sec:understanding} determines \emph{which} token updates are
modified. In particular, naturally contracting tokens that coexist in the
same risky segment are not directly gated.
We do not gate the positive-advantage amplifying region
$\{A_i>0,\ \delta_{i,t}>0\}$: the directional imbalance identified in
Section~\ref{sec:understanding} concentrates in the negative-advantage
region, and attenuating probability increases on reward-positive
responses would directly suppress the learning signal.

The weighted policy loss is
\begin{equation}
    \mathcal{L}^{\mathrm{TRIAGE}}_{\mathrm{PG},i}
    =
    \frac{
        \sum_t
        m_{i,t} w_{i,t}
        \ell^{\mathrm{base}}_{i,t}
    }{
        \sum_t
        m_{i,t} w_{i,t}
    },
    \label{eq:TRIAGE_pg_loss}
\end{equation}
where $\ell^{\mathrm{base}}_{i,t}$ denotes the existing
importance-corrected token loss. Relative to the original response mean,
this induces
\begin{equation}
    \alpha_{i,t}
    =
    \frac{w_{i,t}}{\bar{w}_i},
    \qquad
    \bar{w}_i
    =
    \frac{\sum_t m_{i,t}w_{i,t}}
         {\sum_t m_{i,t}}.
    \label{eq:effective_alpha}
\end{equation}
The normalization therefore reallocates relative update mass away from the
targeted amplifying subset rather than uniformly weakening the entire
negative-advantage response, directly modifying the balance between
$B_{\mathrm{amp}}$ and $B_{\mathrm{con}}$ identified in
Section~\ref{sec:understanding}.

\subsection{Repair on Positive-Advantage Responses}
\label{sec:TRIAGE_repair}

Direction-selective gating controls the allocation of existing policy
gradients, but it is multiplicative: it can attenuate an amplifying
update toward zero, yet cannot reverse its sign or directly reduce the
remaining learner--sampler gap. We therefore add a repair
mechanism on positive-advantage responses. On these responses, the
policy gradient on negative-gap tokens is already mismatch-contracting,
so raising sampled-token log-probabilities reinforces a motion that
agrees with reward-driven learning; applying the same correction to
negative-advantage responses would instead oppose their policy
gradients, which is why that side is handled by gating alone.

Let $R$ denote a repair segment of response $i(R)$, with
$n_R=\sum_{t\in R}m_{i(R),t}$ valid tokens and mean gap
$\bar{\delta}_R$. We define
\begin{equation}
    q_R
    =
    [\tau_{\mathrm{rep}}-\bar{\delta}_R]_+,
    \qquad
    \phi_\beta(q)
    =
    \beta^2
    \left[
        \sqrt{1+(q/\beta)^2}-1
    \right],
    \label{eq:repair_penalty}
\end{equation}
where $q_R$ is the one-sided violation of the repair boundary and
$\phi_\beta$ is a pseudo-Huber penalty.

Gating is active throughout training, whereas repair is enabled from a
fixed update $k_{\mathrm{rep}}$ onward, with
$\lambda_k=\lambda_{\mathrm{rep}}\mathbf{1}[k\geq k_{\mathrm{rep}}]$.
All operating thresholds and the repair schedule are fixed before each
run. At update $k$, the objective is
\begin{equation}
    \mathcal{L}_k
    =
    \mathcal{L}_{\mathrm{PG}}^{\mathrm{TRIAGE}}
    +
    \frac{\lambda_k}{N_{\mathrm{rep}}}
    \sum_{R\in\mathcal{S}_{\mathrm{rep}}^{\mathrm{valid}}}
    \mathbf{1}[A_{i(R)}>0]\,
    \phi_\beta(q_R),
    \label{eq:total_TRIAGE_loss}
\end{equation}
Under the ascent-form notation of Equation~\ref{eq:TRIAGE_update}, one
gradient-descent step on the repair objective contributes the additive
coefficient
\begin{equation}
    \psi_{i,t}
    =
    \frac{\lambda_k}{N_{\mathrm{rep}}}
    \frac{m_{i,t}}{n_R}\,
    \phi_\beta'(q_R)\,
    \mathbf{1}[A_i>0,\ q_R>0].
    \label{eq:repair_psi}
\end{equation}
All valid tokens within an active segment share the same corrective
coefficient. For negative-gap tokens, $\delta_{i,t}\psi_{i,t}<0$, so
the repair term contributes to contraction in the diagonal
decomposition of Section~\ref{sec:understanding}. Positive-gap tokens
may also receive the coefficient: the objective repairs negative
segment-level displacement rather than enforcing contraction for every
token. Appendix~\ref{app:repair_details} derives the positive-advantage
repair coefficient and its bound.

\begin{figure*}[tbp]
    \centering
    \includegraphics[width=\textwidth]{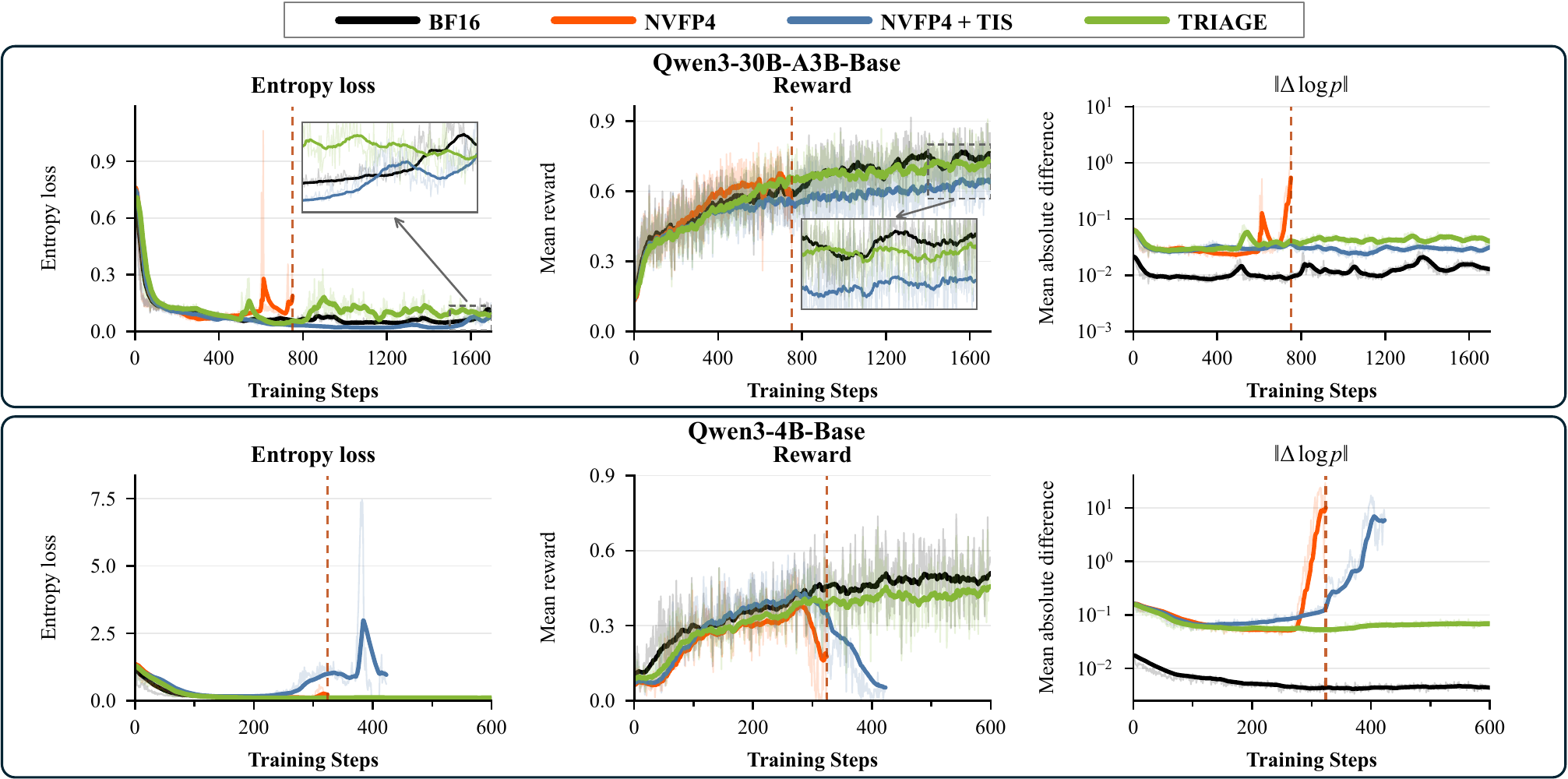}
    \caption{Training dynamics on Qwen3-30B-A3B-Base and Qwen3-4B-Base. 
    From left to right: entropy loss, training reward, and mean absolute train--rollout log-probability
    gap against optimizer steps.}
    \label{fig:main_curves}
\end{figure*}

\section{Experiments}
\label{sec:experiments}
\subsection{Experimental Setup}
\label{sec:experimental_setup}
\paragraph{Models and task.}
We use Qwen3-30B-A3B-Base as the primary model and Qwen3-4B-Base for
replication and ablation studies.
We perform RL with GRPO on the DAPO dataset. To better understand the
quantization error, we use 20,480 as the max response length. Detailed training settings are
listed in Appendix~\ref{app:training_details}.

\paragraph{Settings.}
We compare four training configurations:
(i) \textbf{BF16}, the baseline;
(ii) \textbf{NVFP4}, native W4A4 execution with Transformer Engine;
(iii) \textbf{NVFP4+TIS}, adding token-level importance
correction~\citep{qiu2026fp8rl};
and (iv) \textbf{NVFP4+TRIAGE}, our full method.
All NVFP4 configurations share the same native execution path: W4A4 kernels
with per-token second-level activation scaling~\citep{chan2026composer} on both
the sampler and the learner forward pass.
TIS uses a truncation threshold of $C=\text{2.0}$.
TRIAGE operating rules and segment definitions are given in
Appendix~\ref{app:gate_details}; quantization and system settings are
described in Appendices~\ref{app:training_details} and~\ref{app:efficiency}.

\paragraph{Evaluation Benchmarks and Metrics.}
We evaluate mathematical reasoning on several widely used benchmarks,
including GSM8K~\citep{cobbe2021training},
MATH-500~\citep{lightman2023lets,hendrycks2021measuring},
AIME 2024/2025, and AMC 2023~\citep{maa_amc}.
We also track training reward, policy entropy, and the mean absolute
train--rollout log-probability gap throughout training.
\subsection{Experimental Results}
\label{sec:main_results}
\paragraph{Training stability.}
Figure~\ref{fig:main_curves} shows that naive NVFP4 becomes
unstable after approximately 300 steps on Qwen3-4B-Base,
with a sharp reward drop and rapid growth in the mean absolute
train--rollout log-probability gap.
TIS delays this failure but does not prevent it.
On Qwen3-30B-A3B-Base, naive NVFP4 also exhibits rapid mismatch
growth after roughly 700 steps.
TIS does not exhibit a comparable divergence over 1,700 steps,
but its reward trajectory remains below BF16 and TRIAGE in
later training.
In contrast, TRIAGE remains stable throughout the evaluated
600- and 1,700-step schedules, approximately doubling the
observed stable training horizon on 4B relative to naive NVFP4.
\paragraph{Benchmark performance.}
Table~\ref{tab:benchmark_eval} shows that TRIAGE achieves
near-BF16 average performance: 58.49\% versus 58.26\% on 4B,
and 70.96\% versus 72.41\% on 30B.
At the shared step-300 checkpoint on 4B, TRIAGE already
outperforms naive NVFP4 and NVFP4+TIS by 5.93 and 2.27
percentage points, respectively, showing that its gains
are not solely due to longer training.
Although NVFP4+TIS does not collapse over the full 1700-step training
horizon, it still scores below the pre-collapse naive NVFP4
checkpoint at step 700 (63.18\% versus 65.87\%). 

These observations are consistent with the directional and
localized mismatch analysis in Section~\ref{sec:understanding}:
the global mean does not capture how mismatch interacts with
policy updates or where it concentrates within responses, 
supporting the importance of selective control of mismatch
amplification.
\begin{table*}[tbp]
    \centering
    \caption{
    Performance across five benchmarks.
    Step denotes the evaluated training checkpoint.
    Stable runs are evaluated at the final training step, while runs marked
    with $\dagger$ collapse before the planned training budget and are
    evaluated at the last saved checkpoint before collapse.
    Best result among NVFP4 settings for each model is shown in \textbf{bold}.
    }
    \label{tab:benchmark_eval}
    \footnotesize
    \setlength{\tabcolsep}{6pt}
    \begin{tabular}{lllcccccc}
        \toprule
        Model & Setting & Step
        & AIME24 & AIME25 & MATH500 & AMC23 & GSM8K & \textbf{Avg} \\
        \midrule

        \multirow{4}{*}{30B}
        & BF16        & 1700 & 53.89\% & 36.67\% & 92.65\% & 88.04\% & 90.81\% & \textbf{72.41\%} \\
        \cmidrule(lr){2-9}
        & NVFP4$^\dagger$    & 700 & 44.44\% & 30.00\% & 88.80\% & 77.17\% & 88.95\% & \textbf{65.87\%} \\
        & NVFP4+TIS      & 1700 & 36.11\% & 27.22\% & 88.30\% & 69.02\% & 95.26\% & \textbf{63.18\%} \\
        & NVFP4+TRIAGE   & 1700 & \textcolor{nv}{51.67\%} & \textcolor{nv}{\textbf{38.33\%}} & \textcolor{nv}{91.60\%} & \textcolor{nv}{77.72\%} & \textcolor{nv}{\textbf{95.47\%}} & \textcolor{nv}{\textbf{70.96\%}} \\

        \midrule

        \multirow{5}{*}{4B}
        & BF16        & 600 & 24.44\% & 23.33\% & 84.1\% & 67.40\% & 92.04\% & \textbf{58.26\%} \\
        \cmidrule(lr){2-9}
        & NVFP4$^\dagger$        & 300 & 8.33\% & 8.33\% & 78.4\% & 50.00\% & 90.45\% & \textbf{47.10\%} \\
        & NVFP4+TIS$^\dagger$    & 300 & 16.11\% & 18.33\% & 80.90\% & 53.75\% & 84.72\% & \textbf{50.76\%} \\
        & NVFP4+TRIAGE & 300 & 19.44\% & 17.22\% & 82.00\% & 56.52\% & 89.99\% & \textbf{53.03\%} \\
        & NVFP4+TRIAGE & 600 & \textcolor{nv}{\textbf{27.22\%}} & \textcolor{nv}{20.00\%} & \textcolor{nv}{\textbf{85.80\%}} & \textcolor{nv}{67.39\%} & \textcolor{nv}{\textbf{92.04\%}} & \textcolor{nv}{\textbf{58.49\%}} \\
        \bottomrule
    \end{tabular}
\end{table*}
\begin{table}[tbp]
    \centering
    \caption{
        Natural-EOS generation from the base checkpoint. 
        Values are means over ten post-warm-up iterations at two global
        batch sizes. Initialization, evaluation, and checkpoint saving
        are excluded. Parentheses report speedups over BF16.
    }
    \label{tab:efficiency}
    \footnotesize
    \begin{tabular}{lcccc}
        \toprule
        & \multicolumn{2}{c}{End-to-end cost (s/iter) $\downarrow$} & \multicolumn{2}{c}{Rollout throughput (token/s) $\uparrow$} \\
        \cmidrule(lr){2-5}
        Setting & Batch 256 & Batch 512 & Batch 256 & Batch 512 \\
        \midrule
        BF16
            & 208.2 (1.00$\times$)
            & 254.5 (1.00$\times$)
            & 2977 (1.00$\times$)
            & 3798 (1.00$\times$) \\
        NVFP4
            & 169.4 (1.23$\times$)
            & 191.0 (1.33$\times$)
            & 6201 (2.08$\times$)
            & 8942 (2.35$\times$)\\
        NVFP4+TRIAGE
            & \textcolor{nv}{172.1 (1.21$\times$)}
            & \textcolor{nv}{195.8 (1.30$\times$)}
            & \textcolor{nv}{6745 (2.27$\times$)}
            & \textcolor{nv}{8717 (2.30$\times$)}\\
        \bottomrule
    \end{tabular}
\end{table}
\subsection{Rollout Throughput and Learner Overhead}
\label{sec:efficiency}
\paragraph{Rollout throughput.}
With TRIAGE, native NVFP4 achieves 6,745 and 8,717 output tokens/s
at global batch sizes 256 and 512, respectively, corresponding to
$2.27\times$ and $2.30\times$ the BF16 throughput.
TRIAGE thus retains the more-than-twofold rollout throughput
advantage observed for native NVFP4 in these runs.
\paragraph{End-to-end RL efficiency.}
Iteration timing includes rollout, learner updates, weight
conversion and synchronization, and orchestration.
TRIAGE reduces average iteration time from 208.2 to 172.1\,s
at batch size 256 and from 254.5 to 195.8\,s at batch size 512,
yielding $1.21\times$ and $1.30\times$ end-to-end speedups,
respectively.
\paragraph{TRIAGE overhead.}
We separately isolate TRIAGE's incremental learner cost on eight
B300 GPUs using five paired repetitions of the same frozen batch.
Enabling gating and repair yields an observed median
learner-update overhead of 0.84\% and a median absolute increase
of 0.225\,s.

Detailed throughput settings, environment settings, independent configuration-selection results,
stress-tests results, and complete timing distributions are provided in Appendix~\ref{app:efficiency}.
\subsection{Ablation Studies}
We perform component ablations on Qwen3-4B-Base. All variants branch from the same full-TRIAGE 
checkpoint at step 300.
Full TRIAGE achieves the highest reward.
Removing repair keeps the mean log-probability gap low but leaves
the advantage-driven update imbalance uncorrected, reducing the reward.
TIS alone remains stable when continued from the TRIAGE checkpoint,
unlike its failure when used from the beginning in Figure~\ref{fig:main_curves}.
This contrast reflects the stage-dependent mismatch dynamics
identified in Section~\ref{sec:understanding}.
Naive NVFP4 also allows mismatch to grow sharply, leading
to a late-stage collapse.
\label{sec:ablations}
\begin{figure*}[tbp]
    \centering
    \includegraphics[width=\textwidth]{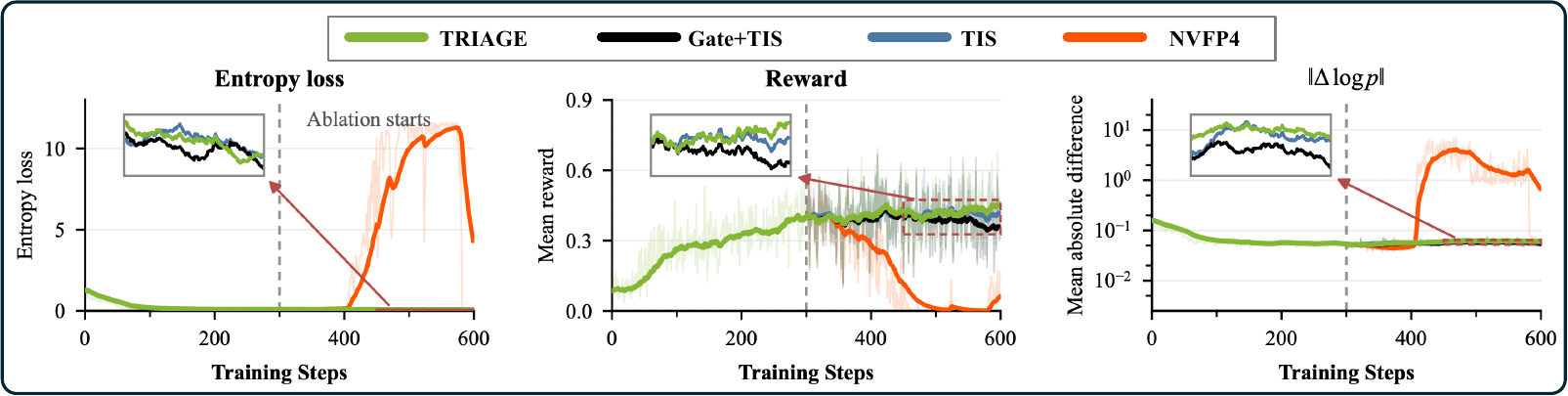}
    \caption{Ablation on Qwen3-4B-Base.
    \textbf{Full TRIAGE};
    \textbf{Gate+TIS} removes repair; \textbf{TIS} additionally removes gating;
    \textbf{NVFP4} removes all three components.
    The dashed line marks the branch point.}
    \label{fig:ablation}
\end{figure*}

\section{Related Work}
\paragraph{Low-precision training.}
Native low-precision training requires careful control of quantization
error. NVIDIA's NVFP4 pretraining recipe~\citep{nvidia2025nvfp4,xie2025infir}
and Quartet II~\citep{panferov2026quartetii} address this through
quantization design and improved gradient estimation.
For RL, QeRL~\citep{huang2025qerl} combines NVFP4 with LoRA and
adaptive quantization noise. FP8-RL~\citep{qiu2026fp8rl} stabilizes
quantized rollout with importance-based corrections. We focus on the optimization dynamics of the residual
learner--sampler mismatch under native NVFP4 W4A4 execution.

\paragraph{Training--inference alignment.}
Implementation differences can destabilize RL even with synchronized
model weights~\citep{zhong2026diagnosing}.
QaRL~\citep{gu2026qarl} aligns the learner with quantized rollout
through low-bit forward computation and introduces sequence-level
dual clipping for negative samples.
QUADS~\citep{zhuge2026quads} combines weight-only fake quantization
on the learner with rollout-side activation compensation.
TRIAGE instead controls how residual mismatch evolves under policy
updates, modifying the optimization objective while retaining native
W4A4 forward execution on both the learner and sampler.

\paragraph{Directional control.}
ACRL~\citep{fan2026acrl} uses advantage-guided, four-quadrant
probability adjustment, with response-level discrepancy relative to
a reference controlling token-wise reweighting.
TRIAGE derives a local mismatch-energy decomposition and identifies
directional asymmetry and localized negative tails in native NVFP4
runs.

\section{Conclusion}
We presented TRIAGE, a direction-aware stabilization method for native
NVFP4 RL. Our analysis distinguishes locally amplifying and contracting
policy-gradient contributions and identifies directional imbalance
and segment-localized negative tails before global mismatch growth.
TRIAGE translates these observations into selective gating,
within-response rebalancing, and bounded repair, while preserving
native W4A4 forward execution on both learner and sampler.
On Qwen3-4B-Base and Qwen3-30B-A3B-Base, TRIAGE sustains stable
optimization over the evaluated horizons and achieves near-BF16
average performance across five mathematical reasoning benchmarks.
In our measured configurations, native NVFP4 with TRIAGE achieves
up to $2.30\times$ rollout throughput and $1.30\times$ end-to-end
speedups over BF16.
These results support controlling how mismatch evolves under policy
updates, rather than uniformly minimizing its magnitude.

\section{Limitations and Future Work}
Our study focuses on establishing the importance of
\emph{direction-aware} stabilization for low-precision RL.
While truncated importance sampling (TIS) can stabilize the
absolute mismatch gap, our 30B experiments show that completing the
training trajectory does not necessarily translate into strong
downstream performance. In contrast, TRIAGE explicitly accounts for
whether learner--sampler mismatch is locally corrected or amplified by
the policy update. Nevertheless, our empirical study remains limited in
scale. Due to computational constraints, we do not evaluate models at
the 100B+ scale or substantially longer training horizons. Our main
runs are terminated after the optimization trajectories exhibit clear
convergence or stable late-stage behavior. Evaluating whether the same
directional dynamics persist over substantially longer training and on
larger sparse models is therefore an important direction for future
work.

Our current experiments also focus on synchronous RL. The directional
criterion underlying TRIAGE is not inherently tied to synchronous
sampling, and may be particularly relevant to asynchronous RL, where
policy staleness introduces an additional source of learner--sampler
mismatch. However, we have not experimentally validated TRIAGE in this
setting. Extending the analysis to jointly characterize quantization
mismatch and asynchronous policy staleness is an interesting direction
for future study.

Finally, we believe that the end-to-end efficiency of native NVFP4 RL remains below
the acceleration potential of Blackwell and Vera Rubin Tensor Core computation.
Although NVFP4 substantially accelerates rollout generation in our
experiments, the overall RL iteration speedup is approximately
$1.3\times$. A growing fraction of runtime is instead spent on
operations outside the accelerated GEMMs, including parameter
synchronization between the trainer and sampler and quantization-related
reductions and data transformations such as Amax computation. Future
systems could therefore obtain larger gains by co-designing TRIAGE with
more tightly integrated low-precision kernels and RL frameworks,
including more efficient parameter synchronization and fused
quantization paths. On the algorithmic side, our directional analysis
also suggests a broader design space beyond the current repair rule:
rather than treating all mismatch-amplifying tokens uniformly, future
work could develop finer-grained repair mechanisms that adapt to the
magnitude, locality, and persistence of directional mismatch.

\bibliography{iclr2027_conference}
\bibliographystyle{iclr2027_conference}

\appendix
\clearpage
\section*{Acknowledge}
This work was supported by the Hong Kong RGC (TRS: T41-517/25-N; GRF: 15228325), ITC RAISe+ (RAI/24/1/086A), and the Research Institute for Generative Artificial Intelligence at PolyU. 
\section*{Appendix Overview}
The appendices provide theoretical derivations, additional
empirical analyses, implementation details, and experimental
protocols supporting the main text. The appendix is organized
as follows.

\begin{itemize}[leftmargin=1.6em]
    \item \textbf{Appendix~\ref{app:mismatch_derivation}: Derivation of Learner--Sampler Mismatch Dynamics.}
    Extends the mismatch-energy analysis in Section~3.1
    to preconditioned policy updates, explicitly accounting
    for cross-token interactions and optimizer residuals.
    \begin{itemize}[label=--, leftmargin=1.6em]
        \item Section~\ref{app:fixed_sampler_setup}: Fixed-Sampler Setup and Update Coefficients.
        \item Section~\ref{app:preconditioned_gap}: Preconditioned Gap and Energy Dynamics.
        \item Section~\ref{app:diag_contributions}: Diagonal Contributions and Their Scope.
        \item Section~\ref{app:amp_con_masses}: Amplifying and Contracting Masses.
    \end{itemize}

    \item \textbf{Appendix~\ref{app:characterization}: Characterization Protocol and Additional Analyses.}
    Details the mismatch characterization protocol and
    analysis windows, examines the empirical validity of
    the directional risk indicator, and provides additional
    localization analyses, full training trajectories,
    and sensitivity studies.
    \begin{itemize}[label=--, leftmargin=1.6em]
        \item Section~\ref{app:analysis_setup:runs}: Models, Training Setup, and Evaluation.
        \item Section~\ref{app:analysis_setup:estimands}: Mismatch Estimands and Analysis Windows.
        \item Section~\ref{app:directional_validity}: Empirical Scope of the Directional Risk Indicator.
        \item Section~\ref{app:segment_analysis}: Additional Analysis of Localized Mismatch.
        \item Section~\ref{app:full_trajectory_context}: Full-Resolution Characterization Trajectories.
        \item Section~\ref{app:aggregation_sensitivity}: Sensitivity to Aggregation Choices.
    \end{itemize}

    \item \textbf{Appendix~\ref{app:gate_details}: TRIAGE Specification and Derivations.}
    Specifies the segment-level gating rules, derives
    the effective update coefficients and bounded repair
    objective, and provides implementation details
    and operating configurations.
    \begin{itemize}[label=--, leftmargin=1.6em]
        \item Section~\ref{app:gate_operating_rule}: Segment Construction and the Gate Operating Rule.
        \item Section~\ref{app:gate_reduction}: From the Weighted Loss to the Effective Update.
        \item Section~\ref{app:repair_details}: Repair Objective and Additive Coefficient.
        \item Section~\ref{app:triage_implementation}: Implementation and Configuration.
    \end{itemize}

    \item \textbf{Appendix~\ref{app:training_details}: Training and Evaluation Protocols.}
    Describes the main training comparisons, checkpoint
    selection, downstream benchmarks, and evaluation
    metrics used in Section~5.
    \begin{itemize}[label=--, leftmargin=1.6em]
        \item Section~\ref{app:main_training}: Main Training Comparisons.
        \item Section~\ref{app:evaluation_details}: Downstream Evaluation.
    \end{itemize}

    \item \textbf{Appendix~\ref{app:efficiency}: Efficiency and Profiling.}
    Reports the measurement protocols for rollout
    throughput, end-to-end RL efficiency, and learner
    overhead, together with fixed-output sampler
    evaluations and long-tail workload stress tests.
    \begin{itemize}[label=--, leftmargin=1.6em]
        \item Section~\ref{app:natural_rl_efficiency}: Natural-Generation RL Measurements.
        \item Section~\ref{app:learner_overhead}: Paired Learner-Overhead Measurement.
        \item Section~\ref{app:fixed_output_sampler}: Fixed-Output Sampler Evaluation.
        \item Section~\ref{app:longtail_workload}: Controlled Long-Tail Workload.
    \end{itemize}
\end{itemize}
\section{Derivation of Learner--Sampler Mismatch Dynamics}
\label{app:mismatch_derivation}

This appendix derives the local mismatch dynamics in
Section~\ref{sec:mismatch_dynamics}, including a positive-semidefinite
preconditioner and explicit optimizer residuals.

\subsection{Fixed-Sampler Setup and Update Coefficients}\label{app:fixed_sampler_setup}

Let $p\equiv(i,t)$ index a valid rollout token, with sampled token $a_p$
and prefix $s_p$ corresponding to $y_{i,t}$ and $(x,y_{i,<t})$ in the main
text. Define
\begin{equation}
    z_p(\theta)=\log\pi_\theta(a_p\mid s_p),\qquad
    b_p=\log\pi_{\mathrm{s}}(a_p\mid s_p),\qquad
    \delta_p(\theta)=z_p(\theta)-b_p.
    \label{eq:app_delta}
\end{equation}
During one learner update, the trajectories, prefixes, masks, advantages,
and recorded sampler log-probabilities are fixed. Hence
$\nabla_\theta\delta_p=\nabla_\theta z_p$. The Taylor expansion below
assumes locally bounded first and second derivatives of the
log-probability map. All gradients are evaluated at the pre-update
parameters $\theta_k$.

Write the current policy-gradient signal as
\begin{equation}
    u_k=\sum_p g_p\nabla_\theta z_p.
    \label{eq:app_policy_signal}
\end{equation}
Here $g_p$ includes the loss-reduction factors. The main-text directional
discussion suppresses these positive factors when writing $g_{i,t}=A_i$;
we retain them here to specify coefficient magnitudes. For the vanilla GRPO
response mean at the start of a single-update iteration, where the
learner/old-learner ratio is one and clipping is inactive,
\begin{equation}
    g_{i,t}^{\mathrm{eff}}=\gamma_k\frac{A_i}{n_i},
    \qquad
    n_i=\sum_t m_{i,t}>0,
    \quad m_{i,t}\in\{0,1\},
    \label{app:eq:effective_g}
\end{equation}
on valid tokens, with zero coefficient on masked tokens. Here $\gamma_k>0$
is the common outer averaging factor for the update; for the single-prompt
group objective in Section~\ref{sec:preliminaries}, $\gamma_k=1/G$. Thus
$\operatorname{sign}(g_{i,t}^{\mathrm{eff}})=\operatorname{sign}(A_i)$
for valid tokens. A positive batch-wide reduction constant preserves this
sign, whereas the response-dependent factor $1/n_i$ must be retained when
comparing coefficient magnitudes. Importance weighting, clipping, masking,
and TRIAGE are represented by their actual effective coefficients $g_p$;
a token whose coefficient is zero has no diagonal update contribution.

\subsection{Preconditioned Gap and Energy Dynamics}\label{app:preconditioned_gap}

Express the local optimizer update as
\begin{equation}
    \Delta\theta_k
    =\theta_{k+1}-\theta_k
    =\eta\mathcal P_k u_k+r_k^{\mathrm{opt}},
    \qquad \eta>0,\quad\mathcal P_k\succeq0,
    \label{eq:app_generic_optimizer}
\end{equation}
where $\mathcal P_k$ is symmetric. The residual $r_k^{\mathrm{opt}}$
contains components outside the preconditioned current-gradient signal,
such as history-dependent momentum and decoupled weight decay. This
decomposition accommodates the local current-gradient component of
AdamW-style updates while keeping those residual contributions explicit.

Let $J$ have rows $\nabla_\theta z_p^\top$. Define
\begin{equation}
    K^{(\mathcal P)}=J\mathcal P_kJ^\top,\qquad
    K^{(\mathcal P)}_{p,q}
    =\nabla_\theta z_p^\top\mathcal P_k\nabla_\theta z_q,
    \qquad
    \mathbf r_k^z=Jr_k^{\mathrm{opt}}.
    \label{eq:app_gram_matrix}
\end{equation}
The matrix $K^{(\mathcal P)}$ is positive semidefinite, although its
off-diagonal entries can have either sign. Applying Taylor expansion with
the sampler fixed gives
\begin{equation}
    \Delta\boldsymbol\delta
    =J\Delta\theta_k+\mathcal O(\|\Delta\theta_k\|_2^2)
    =\eta K^{(\mathcal P)}\mathbf g+\mathbf r_k^z
      +\mathcal O(\|\Delta\theta_k\|_2^2).
    \label{eq:app_delta_vector}
\end{equation}

For $D(\theta)=\frac12\|\boldsymbol\delta(\theta)\|_2^2$, the exact
energy difference is
\begin{equation}
    \Delta D=\boldsymbol\delta^\top\Delta\boldsymbol\delta
             +\frac12\|\Delta\boldsymbol\delta\|_2^2.
    \label{eq:app_energy_expansion}
\end{equation}
Since $\Delta\boldsymbol\delta=\mathcal O(\|\Delta\theta_k\|_2)$,
substitution yields
\begin{equation}
    \Delta D
    =\eta\boldsymbol\delta^\top K^{(\mathcal P)}\mathbf g
      +R_{\mathrm{opt}}+\mathcal O(\|\Delta\theta_k\|_2^2),
    \qquad
    R_{\mathrm{opt}}=\boldsymbol\delta^\top\mathbf r_k^z.
    \label{eq:app_energy_first_order}
\end{equation}
Setting $\mathcal P_k=I$ and $r_k^{\mathrm{opt}}=0$ recovers
$K=JJ^\top$ and the $\mathcal O(\eta^2)$ expansion in
Equation~\ref{eq:mismatch_dynamics} of the main text.

\subsection{Diagonal Contributions and Their Scope}\label{app:diag_contributions}

Separate the diagonal and cross-token terms by defining
\begin{equation}
    \kappa_p=K^{(\mathcal P)}_{p,p}
      =\nabla_\theta z_p^\top\mathcal P_k\nabla_\theta z_p\geq0,
    \qquad
    R_{\mathrm{cross}}
      =\sum_p\sum_{q\ne p}\delta_p K^{(\mathcal P)}_{p,q}g_q.
    \label{eq:app_diag_cross}
\end{equation}
Then
\begin{equation}
    \Delta D
    =\eta\left(\sum_p\kappa_p\delta_p g_p+R_{\mathrm{cross}}\right)
      +R_{\mathrm{opt}}+\mathcal O(\|\Delta\theta_k\|_2^2).
    \label{eq:app_energy_decomposition}
\end{equation}
For the token energy $D_p=\frac12\delta_p^2$, the diagonal contribution
of the current gradient is
\begin{equation}
    \Delta D_p^{\mathrm{diag}}=\eta\kappa_p\delta_p g_p.
    \label{eq:app_local_energy}
\end{equation}
Whenever $\kappa_p>0$, its sign equals $\operatorname{sign}(\delta_p g_p)$:
positive values amplify the existing gap and negative values contract it
within this diagonal contribution. For the vanilla coefficients in
Equation~\ref{app:eq:effective_g}, this is the sign of
$A_i\delta_{i,t}$, giving the two amplifying regions in
Equation~\ref{eq:amplifying_regions}.

The realized motion of the same token also includes the remaining terms:
\begin{equation}
    \delta_p\Delta\delta_p
    =\eta\kappa_p\delta_p g_p
      +\eta\delta_p\sum_{q\ne p}K^{(\mathcal P)}_{p,q}g_q
      +\delta_p r_{k,p}^z
      +\mathcal O(\|\Delta\theta_k\|_2^2).
    \label{eq:app_realized_token_motion}
\end{equation}
Cross-token coupling, optimizer residuals, and higher-order effects can
therefore change the realized sign of $\delta_p\Delta\delta_p$. The
criterion $\delta_p g_p$ isolates the token's own current-gradient
contribution; Appendix~\ref{app:directional_validity} measures its
agreement with realized motion in the characterization runs.

\subsection{Amplifying and Contracting Masses}\label{app:amp_con_masses}

Using $[x]_+=\max(x,0)$, define
\begin{equation}
    B_{\mathrm{amp}}=\sum_p\kappa_p[\delta_p g_p]_+,
    \qquad
    B_{\mathrm{con}}=\sum_p\kappa_p[-\delta_p g_p]_+.
    \label{eq:app_amp_con}
\end{equation}
The identity $x=[x]_+-[-x]_+$ gives
\begin{equation}
    \Delta D
    =\eta\left(B_{\mathrm{amp}}-B_{\mathrm{con}}+R_{\mathrm{cross}}\right)
      +R_{\mathrm{opt}}+\mathcal O(\|\Delta\theta_k\|_2^2).
    \label{eq:app_amp_con_balance}
\end{equation}
Thus $B_{\mathrm{amp}}$ and $B_{\mathrm{con}}$ quantify the two diagonal
contributions, with the optimizer and cross-token effects retained. The
advantage-weighted token masses used in Section~\ref{sec:directional_asymmetry}
are tractable empirical summaries of the amplifying regions; they do not
estimate $B_{\mathrm{amp}}$ or $B_{\mathrm{con}}$, which also depend on
gradient geometry and gap magnitude.

\section{Characterization Protocol and Additional Analyses}
\label{app:characterization}

This appendix specifies the characterization runs, mismatch estimands, and
analysis windows behind Section~\ref{sec:understanding}. It then evaluates
when the diagonal indicator agrees with realized token motion and reports
the complete localization statistics and segment-length sensitivity.

\subsection{Models, Training Setup, and Evaluation}
\label{app:analysis_setup:runs}

\paragraph{Models and training setup.}
We characterize learner--sampler mismatch on
Qwen3-4B-Base and Qwen3-30B-A3B-Base.
Both runs use \textsc{DAPO-Math-17K}, containing 17,398 training
prompts, with learner seed 1234 and rollout/data seed 42.
Each rollout samples 16 prompts with 16 responses per prompt,
giving a global response batch size of 256, followed by one
optimizer update. Sampler weights are refreshed after every
learner update.

Training rollouts use temperature $1.0$, top-$p$ $1.0$, and
top-$k=-1$. The maximum completion length is 16,384 tokens for
Qwen3-4B-Base and 20,480 tokens for Qwen3-30B-A3B-Base.
We use Adam with a constant learning rate of $10^{-6}$ and no
warm-up, $\beta_1=0.9$, $\beta_2=0.98$, $\epsilon=10^{-8}$,
weight decay $0.1$, and gradient clipping at $1.0$.
Master parameters, accumulated gradients, and optimizer states
are maintained in FP32, while the sampler KV cache remains BF16.

\paragraph{NVFP4 configuration.}
We use native NVFP4 (W4A4) with one-dimensional block scaling
and disable $16\times16$ square weight scaling.
Randomized Hadamard transforms (RHT) are enabled, and stochastic
rounding is used for gradient quantization.
For Qwen3-4B-Base, NVFP4 is applied to the gate, up, and down
projections of every MLP layer, while attention, normalization,
and the tied embedding/output weights remain in BF16.
For Qwen3-30B-A3B-Base, all expert MLP layers are quantized
to NVFP4, while attention, normalization, embeddings, routing
gates, and the independent LM head remain in BF16.
Routing replay is enabled for the MoE model.

Table~\ref{tab:analysis_configuration} summarizes the
model-specific settings and the portions of each run used for
the characterization analysis.

\begin{table}[t]
    \centering
    \small
    \setlength{\tabcolsep}{5pt}
    \caption{
    Model-specific configuration of the learner--sampler mismatch
    characterization runs. The characterization horizon denotes
    the updates used in the analysis rather than the configured
    maximum run length.
    }
    \label{tab:analysis_configuration}
    \begin{tabular}{lll}
        \toprule
        Configuration
        & Qwen3-4B-Base
        & Qwen3-30B-A3B-Base \\
        \midrule

        Architecture
        & Dense
        & MoE \\

        NVFP4 scope
        & All MLP Layers
        & All Expert Layers \\

        Max. completion
        & 16,384
        & 20,480 \\

        Routing replay
        & N/A
        & Enabled \\

        Expert parallelism
        & N/A
        & 8 \\

        Characterization horizon
        & 324 updates
        & 724 updates \\

        \bottomrule
    \end{tabular}
\end{table}

\paragraph{Characterization-time evaluation.}
During training, we evaluate AIME-2024 every ten learner updates
using four independently sampled responses per problem.
Decoding uses temperature $0.7$ and top-$p$ $1.0$.
These evaluations are used to track the training trajectories
and determine the analysis windows in
Section~\ref{sec:understanding}; they are distinct from the
final checkpoint evaluation described below.

\paragraph{Independent checkpoint evaluation.}
Final benchmark evaluation is performed independently from the
training sampler using a SGLang service. 
We evaluate AIME-2024, AIME-2025, MATH500, AMC23, and
GSM8K using six sampled responses per
problem.
All benchmarks are decoded with temperature $0.7$, top-$p$
$1.0$, top-$k=-1$, and min-$p=0$, with thinking enabled and a
maximum output length of 20,480 tokens.

The primary metric is the mean sample accuracy
(\texttt{accuracy:mean}), which corresponds to sample-averaged
pass@1 under repeated stochastic decoding.
When additionally reported, pass@$k$ is computed per problem as
\[
    1 -
    \frac{\binom{n-c}{k}}{\binom{n}{k}},
\]
where $n$ is the number of sampled responses and $c$ is the
number scored correct.

\paragraph{Hardware and software.}
All characterization runs use NVIDIA Blackwell B300
GPUs. The training environment uses PyTorch 2.9.1 with CUDA
13.0, Transformer Engine 2.10.0,
miniTransformer 1.0.0, SGLang 0.5.10, and Megatron-LM and
Slime at the revisions released with our artifact.
Exact package revisions, container specifications, launch
commands, and resolved configurations are included in the
released reproducibility artifact.

\subsection{Mismatch Estimands and Analysis Windows}
\label{app:analysis_setup:estimands}

We characterize native-NVFP4 GRPO trajectories of Qwen3-4B-Base and
Qwen3-30B-A3B-Base. The 4B run measures the ordinary learner--sampler gap.
For 30B, learner-side teacher-forced log probabilities are evaluated while
replaying the sampler MoE routing decisions; its measured gap is therefore
a residual mismatch after conditioning on routing:
\begin{equation}
    \delta_{i,t}^{\mathrm{30B,res}}
    =
    z_{i,t}^{\mathrm{learner,route\ replay}}
    -
    b_{i,t},
    \label{app:eq:residual_gap}
\end{equation}
where $b_{i,t}=\log\pi_{\mathrm{s}}(y_{i,t}\mid x,y_{i,<t})$ is the frozen generation-time log-probability and is
reused unchanged in post-update evaluation. Absolute mismatch magnitudes
from the two runs are consequently not treated as directly comparable.

Across successful updates, the logged trajectories contain
$232{,}089{,}183$ valid completion tokens for 4B and
$230{,}713{,}088$ for 30B. Zero-advantage responses account for
$41.27\%$ and $33.16\%$ of these tokens, respectively. Marginal gap
statistics use all valid completion tokens, whereas the
gradient-bearing occupancy reported in the main text excludes
$A_i=0$. The $|A|$-weighted statistics automatically assign zero mass to
such tokens.

For model $m$ with characterization horizon $S_m$, we place analysis
anchors at normalized progress $p\in\{0.25,0.50,0.75,1.00\}$.
The anchor update and its inclusive $25$-update trailing window are
\begin{equation}
    a_m(p)
    =
    \operatorname{round}(pS_m),
    \qquad
    \mathcal{W}_m(p)
    =
    \{a_m(p)-24,\ldots,a_m(p)\}.
    \label{app:eq:trailing_window}
\end{equation}
The four stage-wise analysis windows used in
Section~\ref{sec:directional_asymmetry} are listed in
Table~\ref{tab:characterization_windows}.

\begin{table}[t]
    \centering
    \small
    \setlength{\tabcolsep}{6pt}
    \caption{
    Progress-normalized analysis windows used for learner--sampler mismatch
    characterization. For each model, we place analysis anchors at
    $25\%$, $50\%$, $75\%$, and $100\%$ of the characterization trajectory
    and use the 25-update trailing window ending at each anchor.
    Stage names are descriptive shorthand for these progress-normalized
    anchors rather than independently selected training phases.
    }
    \label{tab:characterization_windows}
    \begin{tabular}{lcccc}
        \toprule
        & Healthy & Drift & Pre-terminal & Terminal \\
        \midrule
        Normalized anchor $p$
        & $25\%$ & $50\%$ & $75\%$ & $100\%$ \\
        \midrule
        Qwen3-4B-Base ($S=324$)
        & 57--81
        & 138--162
        & 219--243
        & 300--324 \\
        Qwen3-30B-A3B-Base ($S=724$)
        & 157--181
        & 338--362
        & 519--543
        & 700--724 \\
        \bottomrule
    \end{tabular}
\end{table}

The collapse trigger is run-specific. For the 4B run, let
$c_s=\mathbf{1}[\mathrm{raw\_reward}_s\leq0.10\ \land\
\mathrm{pass@16}_s\leq0.3125]$
from the logged training metrics at update $s$, the latter computed over
the rollout group of $16$ samples per prompt; the collapse onset is the
first update completing three consecutive violations,
$\min\{s:c_{s-2}=c_{s-1}=c_s=1\}=304$. For the 30B run, the onset is the
first update with a non-finite logged gradient norm, update $724$, which
terminates the trajectory. The two criteria are used only to align stages
within each run and do not assert a shared failure mechanism.

For the 30B terminal window, update 724 contributes its pre-update snapshot
to distributional and segment statistics, but the failed optimizer step is
not included in successful-update token counts. For 4B, the terminal window
extends beyond the first collapse trigger and is used to characterize the
resulting global-mismatch regime rather than to identify the trigger time
itself.
\subsection{Empirical Scope of the Directional Risk Indicator}
\label{app:directional_validity}

The decomposition in Appendix~\ref{app:mismatch_derivation} retains
cross-token interactions and optimizer residuals. We measure directly when
its diagonal sign indicator agrees with the realized one-update motion.

\subsubsection{One-Update Directional Audit}

For a saved anchor $k$, let $\theta_k^{-}$ and $\theta_k^{+}$ denote the
learner parameters before and after one successful optimizer update. We
reuse the identical rollout batch, sampled prefixes, advantages, masks, and
logged sampler log probabilities, and perform teacher-forced learner replay
before and after the update:
\begin{equation}
    \delta_{i,t}^{-}
    =
    z_{i,t}^{-}-b_{i,t},
    \qquad
    \delta_{i,t}^{+}
    =
    z_{i,t}^{+}-b_{i,t}.
\end{equation}
Hence the realized gap motion is
\begin{equation}
    \Delta\delta_{i,t}
    =
    \delta_{i,t}^{+}-\delta_{i,t}^{-}.
    \label{app:eq:realized_delta_change}
\end{equation}

The fixed-sampler identity
\begin{equation}
    \Delta\delta_{i,t}
    =
    z_{i,t}^{+}-z_{i,t}^{-}
\end{equation}
is checked numerically with absolute tolerance $2\times10^{-6}$ and zero
relative tolerance. Replay is executed under \texttt{eval()} and both
attention and hidden-state dropout are disabled. For the 30B model, the MoE
routing cache is reused exactly. We do not snapshot and restore the
kernel-level stochastic-rounding RNG state; the audit therefore measures
the realized implementation-level one-update motion rather than asserting a
bitwise-identical quantization-noise realization before and after the
parameter update.
The reference sampler remains fixed in this audit; it does not measure
the change in execution mismatch after sampler weights are refreshed.

Each probe contains exactly one optimizer step per rollout. An anchor is
retained only when \texttt{optimizer.step()} reports a successful update
and the optimizer scheduler advances. The late anchors are updates
248--252 for 4B and 598--602 for 30B.

For the response-mean reducer, the vanilla token coefficient is
$g_{i,t}^{\mathrm{eff}}=\gamma_k A_i/n_i$
(Eq.~\ref{app:eq:effective_g}). The audit uses
$\widetilde g_{i,t}=A_i/n_i$, retaining the response-length factor and
removing the positive update-common factor $\gamma_k$.
This preserves the diagonal sign and normalized agreement within each
update. Pooled scores across anchors use these normalized coefficients,
without restoring anchor-specific outer factors.
On an eligible token
$j\equiv(i,t)$, define
\begin{equation}
    p_j
    =
    \operatorname{sign}
    (\delta_j^-\widetilde g_j),
    \qquad
    r_j
    =
    \operatorname{sign}
    (\delta_j^-\Delta\delta_j).
\end{equation}
The main audit requires
$|\delta_j^-|>10^{-12}$,
$\widetilde g_j\neq0$, and
$\Delta\delta_j\neq0$; exact zeros are treated as ties and excluded.

For a selected token population $\mathcal{S}$, the unweighted agreement is
\begin{equation}
    \operatorname{Agr}_{\mathrm{unw}}(\mathcal{S})
    =
    \frac{1}{|\mathcal{S}|}
    \sum_{j\in\mathcal{S}}
    \mathbf{1}[p_j=r_j].
    \label{app:eq:agreement_unweighted}
\end{equation}
We additionally weight tokens by the magnitude of their diagonal
directional signal,
\begin{equation}
    u_j
    =
    |\delta_j^-\widetilde g_j|.
\end{equation}
The weighted agreement is
\begin{equation}
    \operatorname{Agr}_{\mathrm{w}}(\mathcal{S})
    =
    \frac{
        \sum_{j\in\mathcal{S}}
        u_j\,\mathbf{1}[p_j=r_j]
    }{
        \sum_{j\in\mathcal{S}}u_j
    }.
    \label{app:eq:agreement_weighted}
\end{equation}
As an outlier-robustness check, we also cap $u_j$ at its $99.9$th
percentile within each reported population before applying
Eq.~\ref{app:eq:agreement_weighted}.

\subsubsection{Agreement as Mismatch Becomes Severe}

Table~\ref{tab:directional_agreement} reports the late-stage audit in the
negative-gap amplifying region
$\mathcal{A}^{-}=\{A_i<0,\delta_{i,t}<0\}$.

\begin{table}[t]
    \centering
    \small
    \setlength{\tabcolsep}{4.5pt}
    \caption{
    Agreement (\%) between the diagonal directional indicator
    $\operatorname{sign}(\delta g^{\mathrm{eff}})$ and realized mismatch
    motion $\operatorname{sign}(\delta\Delta\delta)$ on late-stage
    $\mathcal{A}^{-}$ tokens. ``Weighted'' uses
    $|\delta g^{\mathrm{eff}}|$; ``Capped'' additionally caps this weight at
    the population-specific $99.9$th percentile.
    $\dagger$ denotes low-sample group.
    }
    \label{tab:directional_agreement}
    \begin{tabular}{llrrrr}
        \toprule
        Model & Stratum & Tokens & Unw. & Weighted & Capped \\
        \midrule
        Qwen3-4B-Base
        & All
        & 1,640,752 & 38.88 & 47.73 & 46.81 \\
        Qwen3-4B-Base
        & $|\delta|>0.5$
        & 47,436 & 47.99 & 50.09 & 49.70 \\
        Qwen3-4B-Base
        & $|\delta|>1$
        & 18,822 & 50.43 & 52.34 & 51.99 \\
        Qwen3-4B-Base
        & $|\delta|>3$
        & 1,419 & 59.62 & 63.05 & 63.01 \\
        Qwen3-4B-Base
        & $|\delta|>6$
        & 392$^\dagger$ & 60.20 & 65.75 & 65.73 \\
        \midrule
        Qwen3-30B-A3B-Base
        & All
        & 231,854 & 33.69 & 25.27 & 25.44 \\
        Qwen3-30B-A3B-Base
        & $|\delta|>0.5$
        & 9,228 & 25.88 & 24.07 & 24.07 \\
        Qwen3-30B-A3B-Base
        & $|\delta|>1$
        & 5,887 & 24.27 & 23.27 & 23.35 \\
        Qwen3-30B-A3B-Base
        & $|\delta|>3$
        & 2,026 & 21.08 & 21.86 & 21.95 \\
        Qwen3-30B-A3B-Base
        & $|\delta|>6$
        & 723$^\dagger$ & 19.78 & 22.18 & 22.20 \\
        \bottomrule
    \end{tabular}
\end{table}

For the Qwen3-4B-Base vanilla-mismatch run, the diagonal indicator becomes
substantially more informative as mismatch severity increases. Weighted
agreement rises from $47.73\%$ over all eligible late-stage
$\mathcal{A}^{-}$ tokens to $63.05\%$ for $|\delta|>3$. The latter
population still contains $1{,}419$ tokens across $316$ responses. More
importantly, the weighted agreement for this stratum remains above the
$50\%$ reference at each of the five saved anchors, ranging from
$59.03\%$ to $66.49\%$. Capping the largest $0.1\%$ of weights changes
the pooled result only from $63.05\%$ to $63.01\%$, showing that the trend
is not driven by a handful of extreme-weight tokens. We do not base the
main claim on the $|\delta|>6$ row because of its lower support.

The same token-wise trend does not reproduce for the Qwen3-30B-A3B-Base routing-replay
residual estimand: weighted agreement remains below the $50\%$ reference
and decreases rather than increases with severity. In this conditioned residual setting, the diagonal term in
Equation~\ref{eq:app_amp_con_balance} alone is not a reliable token-wise
predictor. The current measurements do not isolate whether this behavior is
primarily caused by cross-token coupling, routing-conditioned residual
structure, or their interaction. We therefore restrict token-wise
predictiveness claims to the Qwen3-4B-Base vanilla characterization and use the two
models jointly only for the population-level directional and localization
patterns reported in the main text.

\subsection{Additional Analysis of Localized Mismatch}
\label{app:segment_analysis}

This section provides the construction, complete stage-wise statistics, and
sensitivity checks for the segment-level localization analysis in
Section~\ref{sec:localized_mismatch}.

\subsubsection{Segment Construction and Threshold Calibration}

Each response is partitioned independently by its original zero-based
completion-token position $t$: segment $k$ contains valid positions with
$\lfloor t/W\rfloor=k$. Masked positions are excluded from segment
statistics without reindexing the remaining tokens. Empty segments are
omitted, and the final nonempty segment is retained even if it contains
fewer than $W$ valid tokens. No segment crosses a response boundary.
The main analysis uses $W=64$.

Because the localized failure mode of interest is the negative-gap
amplifying region, segment statistics are computed over responses with
$A_i<0$. For a segment $S$ in response $i$, define
\begin{equation}
    f_S^{\mathrm{amp}}(\tau)
    =
    \frac{1}{|S|}
    \sum_{t\in S}
    \mathbf{1}
    [A_i<0,\ \delta_{i,t}<\tau].
    \label{app:eq:amp_segment_fraction}
\end{equation}

The analysis threshold is calibrated once per model from the healthy
reference window:
\begin{equation}
    \tau_{\mathrm{ana}}^{(m)}
    =
    Q_{0.05}
    \left(
        \{
        \delta_{i,t}:
        (i,t)\in\mathcal{T}_{m,\mathrm{healthy}}
        \}
    \right),
    \label{app:eq:analysis_threshold}
\end{equation}
where the quantile is computed over \emph{all valid response tokens} in the
healthy window, rather than only the $A_i<0$ subset. This gives
\begin{equation}
    \tau_{\mathrm{ana}}^{(\mathrm{4B})}
    =
    -0.24208,
    \qquad
    \tau_{\mathrm{ana}}^{(\mathrm{30B})}
    =
    -0.03949.
\end{equation}
A segment is called \emph{tail-heavy} when
\begin{equation}
    h_S
    =
    \mathbf{1}
    \left[
        f_S^{\mathrm{amp}}
        (\tau_{\mathrm{ana}})>0.1
    \right].
    \label{app:eq:tail_heavy}
\end{equation}

This calibration separates the localization question from later global
tail growth: $\tau_{\mathrm{ana}}$ is fixed before drift rather than
re-estimated at every stage.

\subsubsection{Localization Metrics}

Let $\mathcal{S}$ denote all eligible segments pooled within an analysis
window. We rank them by $f_S^{\mathrm{amp}}$ and retain the top
$\lceil0.05|\mathcal{S}|\rceil$ as $\mathcal{S}_{5\%}$.
Ties are resolved by decreasing tail count, followed by canonical
response and segment identifiers.

The amplifying-tail base rate is
\begin{equation}
    r_{\mathrm{amp}}
    =
    \frac{
        \sum_{S\in\mathcal{S}}
        \sum_{t\in S}
        \mathbf{1}[\delta_{i,t}<\tau_{\mathrm{ana}}]
    }{
        \sum_{S\in\mathcal{S}}|S|
    }.
    \label{app:eq:amp_base_rate}
\end{equation}

The fraction of amplifying-tail tokens contained in the top-$5\%$
segments is
\begin{equation}
    C_{\mathrm{tok}}
    =
    \frac{
        \sum_{S\in\mathcal{S}_{5\%}}
        \sum_{t\in S}
        \mathbf{1}[\delta_{i,t}<\tau_{\mathrm{ana}}]
    }{
        \sum_{S\in\mathcal{S}}
        \sum_{t\in S}
        \mathbf{1}[\delta_{i,t}<\tau_{\mathrm{ana}}]
    }.
    \label{app:eq:token_containment}
\end{equation}

The $5\%$ uniform-allocation reference describes equal tail mass per
segment. It is not a random-position null for $C_{\mathrm{tok}}$:
ranking segments by their observed tail fractions raises expected
top-$5\%$ containment even under random placement. We account for this
selection in Section~\ref{app:response_conditioned_localization}.

Tail-heavy segment coverage is
\begin{equation}
    H_{\mathrm{seg}}
    =
    \frac{1}{|\mathcal{S}|}
    \sum_{S\in\mathcal{S}}h_S.
    \label{app:eq:heavy_coverage}
\end{equation}

For adjacency, let $\mathcal{E}$ contain directed pairs
$(S_{i,k},S_{i,k+1})$ for which both segments are nonempty and belong to
the same response. Probabilities over $\mathcal{E}$ weight every eligible
pair equally. We report
\begin{equation}
    P_{\mathrm{adj}}
    =
    \frac{
        \Pr_{\mathcal{E}}(h_{S_{i,k+1}}=1
        \mid
        h_{S_{i,k}}=1)
    }{
        \Pr_{\mathcal{E}}(h_{S_{i,k+1}}=1)
    }.
    \label{app:eq:adjacent_persistence}
\end{equation}
The denominator is the successor-heavy marginal over eligible pairs,
not $H_{\mathrm{seg}}$. Thus $P_{\mathrm{adj}}>1$ indicates association
relative to this pooled marginal. Response-level heterogeneity can also
produce such an association; the response-conditioned controls below
separate this reference from randomized token positions and segment order.

\subsubsection{Stage-Wise Localization with \texorpdfstring{$W=64$}{W=64}}

Table~\ref{tab:segment_w64} reports the full stage-wise evolution of the
core localization statistics.

\begin{table}[t]
    \centering
    \small
    \setlength{\tabcolsep}{3.5pt}
    \caption{
    Stage-wise localization of negative-gap amplifying tokens with
    $W=64$. $r_{\mathrm{amp}}$, $H_{\mathrm{seg}}$,
    and $C_{\mathrm{tok}}$ are percentages.
    $P_{\mathrm{adj}}$ uses the successor-heavy marginal over eligible
    adjacent pairs. The table reports descriptive statistics; matched
    random-position and segment-order controls are shown in
    Figure~\ref{fig:response_conditioned_localization}.
    }
    \label{tab:segment_w64}
    \begin{tabular}{llrrrr}
        \toprule
        Model & Stage
        & $r_{\mathrm{amp}}$
        & $H_{\mathrm{seg}}$
        & $C_{\mathrm{tok}}$
        & $P_{\mathrm{adj}}$ \\
        \midrule
        Qwen3-4B-Base & Healthy
        & 4.90 & 15.08 & 21.56 & 2.65 \\
        Qwen3-4B-Base & Drift
        & 2.60 & 5.43 & 25.74 & 4.22 \\
        Qwen3-4B-Base & Pre-terminal
        & 2.75 & 6.24 & 25.68 & 3.98 \\
        Qwen3-4B-Base & Terminal
        & 31.53 & 48.64 & 15.13 & 1.84 \\
        \midrule
        Qwen3-30B-A3B-Base & Healthy
        & 4.79 & 14.46 & 20.10 & 2.62 \\
        Qwen3-30B-A3B-Base & Drift
        & 3.82 & 10.43 & 22.15 & 3.28 \\
        Qwen3-30B-A3B-Base & Pre-terminal
        & 3.59 & 10.25 & 25.01 & 3.78 \\
        Qwen3-30B-A3B-Base & Terminal
        & 22.91 & 57.02 & 18.09 & 1.45 \\
        \bottomrule
    \end{tabular}
\end{table}

The tail's frequency and its distribution across segments evolve
differently. From the healthy to pre-terminal window, $r_{\mathrm{amp}}$
decreases from $4.90\%$ to $2.75\%$ on 4B and from $4.79\%$ to $3.59\%$
on 30B, while $C_{\mathrm{tok}}$ increases to $25.68\%$ and $25.01\%$.
At the terminal stage, tail-heavy coverage reaches $48.64\%$ and
$57.02\%$, and top-$5\%$ containment decreases. These statistics
distinguish concentrated pre-terminal mismatch from the broader coverage
of the terminal regime.

\begin{figure}[t]
    \centering
    \includegraphics[width=\linewidth]{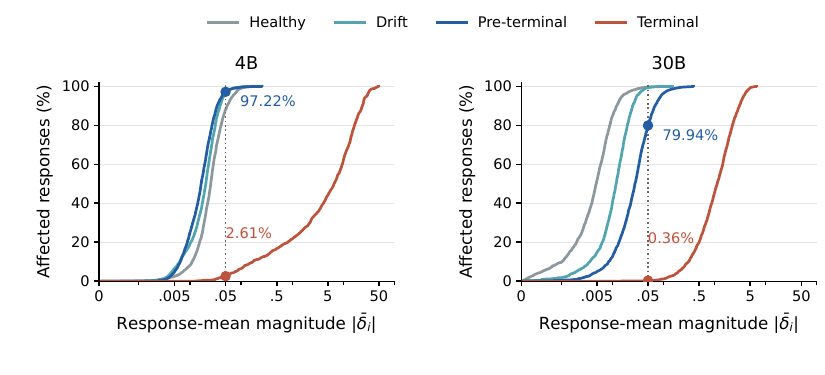}
    \caption{\textbf{Local tails within responses with small mean mismatch.}
    Empirical CDFs of $|\bar{\delta}_i|$ among $A_i<0$ responses with at
    least one tail-heavy $W=64$ segment. All four stage windows are shown
    for each model, with affected responses selected separately in each
    window and weighted equally. The vertical line marks $0.05$; labels give the
    pre-terminal and terminal fractions below this cutoff.}
    \label{fig:affected_response_means}
\end{figure}

Response means can remain small despite these local tails.
Figure~\ref{fig:affected_response_means} shows the full empirical
distribution of $|\bar{\delta}_i|$ among negative-advantage responses
containing at least one tail-heavy segment, with
$\bar{\delta}_i=n_i^{-1}\sum_{t\,\mathrm{valid}}\delta_{i,t}$.
At pre-terminal, $97.22\%$ of affected 4B responses and $79.94\%$ of
affected 30B responses satisfy $|\bar{\delta}_i|<0.05$; the terminal
fractions are $2.61\%$ and $0.36\%$. Replacing the magnitude of the
signed mean with the mean absolute token gap gives pre-terminal fractions
of $52.37\%$ and $50.21\%$ at the same cutoff. Signed cancellation
therefore contributes to dilution, while a local tail can also coexist
with a small mean absolute gap.

\subsubsection{Sensitivity to Segment Length}

We repeat the analysis with $W=128$, retaining the final shorter segment
and using the same model-specific
$\tau_{\mathrm{ana}}$. Table~\ref{tab:segment_w128_sensitivity} summarizes
the healthy-to-pre-terminal transition and the terminal regime.

\begin{table}[t]
    \centering
    \small
    \setlength{\tabcolsep}{4pt}
    \caption{
    Segment-length sensitivity with $W=128$. Entries before and after the
    arrow correspond to the healthy and pre-terminal stages.
    $r_{\mathrm{amp}}$, $C_{\mathrm{tok}}$, and $H_{\mathrm{seg}}$ are
    percentages; $P_{\mathrm{adj}}$ uses the same eligible-pair definition
    as the $W=64$ analysis.
    }
    \label{tab:segment_w128_sensitivity}
    \begin{tabular}{lrrrr}
        \toprule
        Model
        & $r_{\mathrm{amp}}$
        & $C_{\mathrm{tok}}$
        & $P_{\mathrm{adj}}$
        & Terminal $H_{\mathrm{seg}}$ \\
        \midrule
        Qwen3-4B-Base
        & $4.90\!\rightarrow\!2.75$
        & $18.39\!\rightarrow\!21.81$
        & $2.82\!\rightarrow\!5.57$
        & 48.98 \\
        Qwen3-30B-A3B-Base
        & $4.79\!\rightarrow\!3.59$
        & $17.23\!\rightarrow\!21.76$
        & $2.85\!\rightarrow\!4.37$
        & 60.25 \\
        \bottomrule
    \end{tabular}
\end{table}

The sequence-level dilution result is also preserved:
$96.11\%$ and $75.99\%$ of affected 4B and 30B responses satisfy
$|\bar{\delta}_i|<0.05$ at the pre-terminal stage, whereas the
corresponding terminal values fall to $1.86\%$ and $0.23\%$.
At terminal failure, top-$5\%$ token containment also decreases to
$14.83\%$ and $17.63\%$, and persistence approaches
$1.84\times$ and $1.40\times$, respectively. These descriptive trends
are consistent across the two segment lengths. The conditional
randomization analyses below use the default $W=64$ partition.

The fixed severe-gap criterion $\delta<-6$ provides a complementary
severity check. With $W=64$, it yields only $11$, $7$, and $49$
tail-heavy segments in the healthy, drift, and pre-terminal 4B windows,
and $0$, $4$, and $20$ on 30B. This sparsity makes early adjacency ratios
unstable. At terminal, the criterion includes $24.88\%$ and $10.07\%$
of eligible tokens, respectively. We therefore retain the
healthy-calibrated threshold for localization and use $\delta<-6$ to
describe the extreme terminal tail.

\subsubsection{Response-Conditioned Localization Controls}
\label{app:response_conditioned_localization}

\begin{figure}[htbp]
    \centering
    \includegraphics[width=\linewidth]{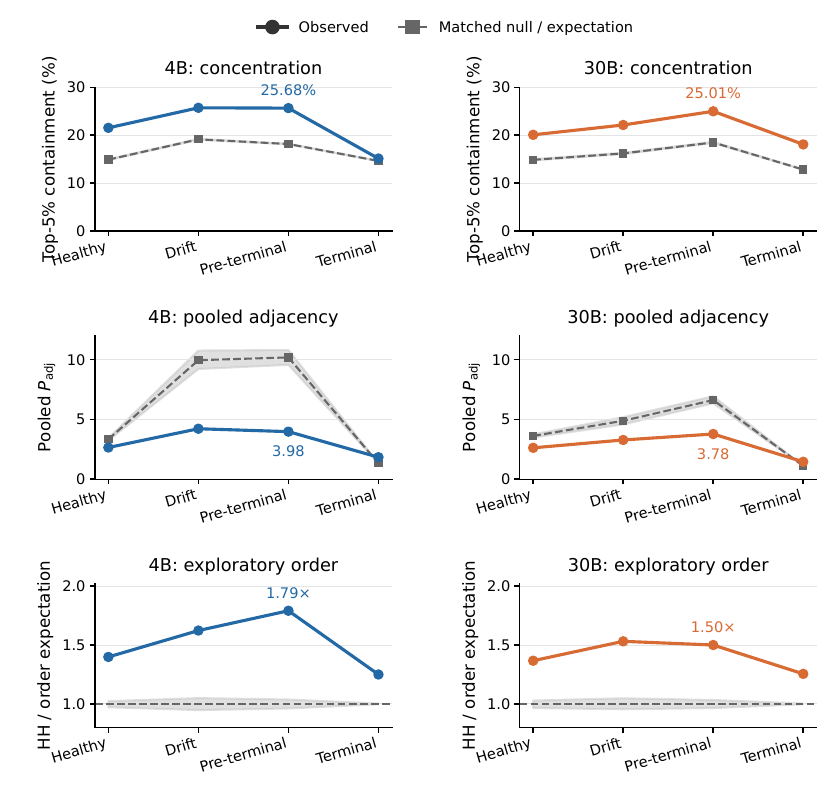}
    \caption{\textbf{Response-conditioned controls for local structure.}
    Top and middle: observed $C_{\mathrm{tok}}$ and $P_{\mathrm{adj}}$
    compared with within-response tail-label shuffling, retaining all
    eligible segments, including short segments. Dashed curves show null
    medians. Bottom: the exploratory segment-order test, restricted to
    full segments and fixing each response's heavy-segment count; the
    dashed line is the analytic random-order expectation normalized to
    one. Shading denotes central $95\%$ conditional randomization
    intervals from 499 draws per window, not across-run confidence
    intervals. Both models and all four windows are retained.}
    \label{fig:response_conditioned_localization}
\end{figure}

To test whether concentration exceeds the effect of response-level tail
rates and top-segment selection, we randomize tail positions within each
response. The analysis retains the same four windows, thresholds, and
$W=64$ segmentation. Figure~\ref{fig:response_conditioned_localization}
reports the original metrics under this control and a separate
exploratory test of segment order.

For every $A_i<0$ response, the primary control fixes valid-token
positions, response length, and total tail-token count, then uniformly
permutes binary tail labels. Each of 499 randomizations recomputes
segment tail fractions, the global top-$5\%$ ranking, and
$P_{\mathrm{adj}}$. Multivariate hypergeometric allocation implements
this permutation exactly while preserving segment capacities, including
short segments. We use randomization seed 20260924 and report central
$95\%$ conditional randomization intervals. These intervals describe
random placement within the observed responses, not variability across
independent training runs.

At pre-terminal, observed $C_{\mathrm{tok}}$ is $25.68\%$ on 4B and
$25.01\%$ on 30B, compared with null medians of $18.17\%$ and
$18.50\%$. The corresponding null intervals are
$[18.11\%,18.23\%]$ and $[18.38\%,18.63\%]$.
The excesses of $7.51$ and $6.51$ percentage points show concentration
beyond response-level tail-rate heterogeneity and rank selection.
Such excess concentration is also present in the healthy windows;
its magnitude does not increase monotonically across both trajectories.

The original $P_{\mathrm{adj}}$ has a different comparison: its
pre-terminal value is $3.98$ versus a token-shuffle median of $10.18$
on 4B, and $3.78$ versus $6.62$ on 30B. Thus its elevation above one
does not establish persistence beyond this response-conditioned null.
Tail-label shuffling changes which segments are heavy, as well as
their arrangement.

After examining this primary control, we conducted an exploratory
segment-order test that conditions on the observed heavy-segment counts.
For response $i$, we retain the original positions of its $m_i$ full
segments and uniformly choose $h_i$ of these positions to carry heavy
labels, preserving the observed heavy-segment count. Partial segments
are excluded, and gaps do not create new adjacency edges. Let $E_i$ be
the set of genuinely adjacent full-segment pairs and $T$ the total
number of heavy--heavy pairs. Its random-order expectation is
\[
    \mathbb{E}_0[T]
    =\sum_{i:m_i\ge2}|E_i|
    \frac{h_i(h_i-1)}{m_i(m_i-1)}.
\]
At pre-terminal, observed counts are $2{,}892$ versus an expectation
of $1{,}613.85$ on 4B, and $1{,}584$ versus $1{,}055.58$ on 30B,
giving enrichments of $1.79\times$ and $1.50\times$.
This comparison identifies non-random ordering conditional on the
observed heavy-segment counts. It is a different statistic from
$P_{\mathrm{adj}}$ and does not exclude within-response position trends
or establish causal propagation along the response.

\subsection{Full-Resolution Characterization Trajectories}
\label{app:full_trajectory_context}

\begin{figure}[htbp]
    \centering
    \includegraphics[width=\linewidth]{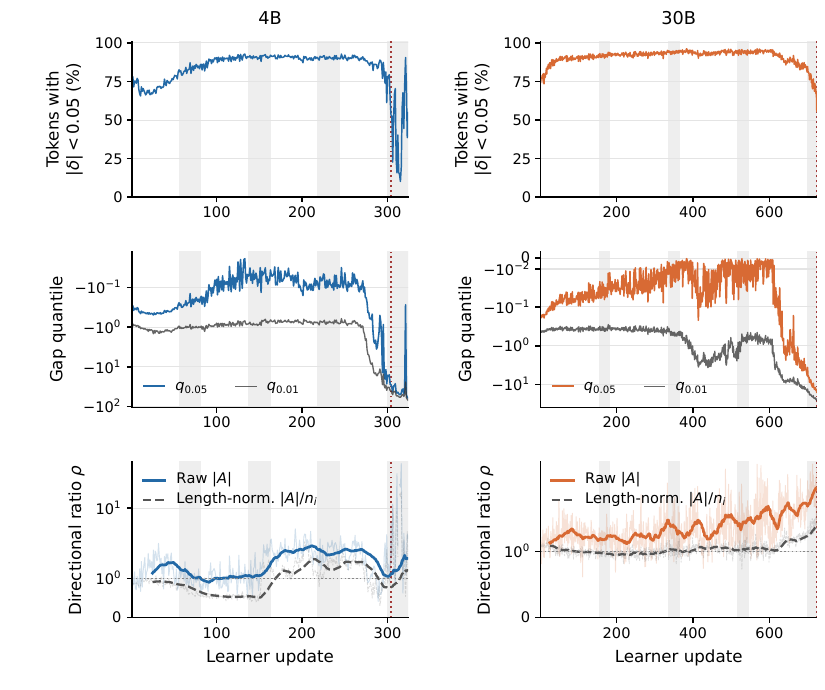}
    \caption{\textbf{Bulk, tail, and directional summaries along the full
    characterization trajectories.} Top: fraction of all valid completion
    tokens with $|\delta|<0.05$ at each update. Middle: exact per-update
    signed gap quantiles $q_{0.05}$ and $q_{0.01}$. Bottom: raw and
    length-normalized $\mathcal{A}^{-}/\mathcal{A}^{+}$ weight ratios,
    with per-update values shown faintly and trailing 25-update pooled
    values in bold. The middle and bottom rows use symmetric-logarithmic
    vertical axes. Gray spans mark the original analysis windows; red
    lines mark the run-specific collapse triggers (4B: 304; 30B: 724).
    The 30B endpoint uses the pre-update snapshot of the failed update.}
    \label{fig:characterization_full_trajectory}
\end{figure}

Figure~\ref{fig:characterization_full_trajectory} places the four
stage-wise summaries of Section~\ref{sec:directional_asymmetry} in the
full saved trajectory: updates 1--324 for 4B and 1--724 for 30B.
For each update, we merge the data-parallel shards and compute the
fraction of valid completion tokens with $|\delta|<0.05$ and the exact
$0.05$ and $0.01$ gap quantiles. These marginal statistics include
zero-advantage tokens. The ordinary 4B gap and routing-conditioned 30B
residual retain their distinct estimands.

The directional panels compare raw $|A_i|$ weights with
response-length-normalized $|A_i|/n_i$ weights over the strict regions
$\mathcal{A}^{-}=\{A_i<0,\delta_{i,t}<0\}$ and
$\mathcal{A}^{+}=\{A_i>0,\delta_{i,t}>0\}$. Faint curves retain
per-update ratios; bold curves pool numerator and denominator weights
over trailing 25-update windows before taking their ratio. They are not
moving averages of per-update ratios. Tokens with $\delta=0$ belong to
neither strict region, and tokens with $A_i=0$ carry zero weight.

The complete trajectories show temporal variation that is lost when
updates are pooled into broad consecutive windows. The directional ratio
is non-monotonic, and length normalization reduces the raw-weighted
imbalance, particularly on 30B. The bulk and tail panels locate these
changes relative to the recorded terminal regime; the four shaded
windows remain descriptive summaries of one characterization run per
model. Section~\ref{app:aggregation_sensitivity} gives the corresponding
stage-wise sensitivity results.

\subsection{Sensitivity to Aggregation Choices}
\label{app:aggregation_sensitivity}
We examine two choices in the directional summaries: the trailing update-window width and the weighting assigned to each valid token. For a region $\mathcal R$, define
\begin{equation*}
 w_q(\mathcal R)=\frac{\sum_{i,t}q_{i,t}\,\mathbf 1[(i,t)\in\mathcal R]}{\sum_{i,t}q_{i,t}},
 \qquad \rho_q=\frac{w_q(\mathcal{A}^{-})}{w_q(\mathcal{A}^{+})}.
\end{equation*}
The stage-based summaries in Table~\ref{tab:sensitivity} retain the
25-update windows of Section~\ref{app:analysis_setup:estimands}, shown
as gray spans in Figure~\ref{fig:characterization_full_trajectory}.
The primary summary uses $q_{i,t}=\lvert A_i\rvert$. The alternative
uses $q_{i,t}=\lvert A_i\rvert/n_i$, where $n_i$ is the valid completion
length of response $i$, to include the response-mean reduction factor.

\begin{table}[htbp]
    \centering
    \caption{\textbf{Sensitivity of directional summaries.}
    The window comparison holds raw advantage weighting fixed.
    The weighting comparison holds each 25-update stage window fixed.
    Masses are percentages of the corresponding total weight.}
    \label{tab:sensitivity}
    \small

    \begin{tabularx}{\linewidth}{@{}lXrrrr@{}}
        \toprule
        \multicolumn{6}{@{}l}{\textbf{(a) Window width: 25 versus 50 trailing updates}}\\
        Model & Stage & $\rho_{25}$ & $\rho_{50}$ 
        & $w_{25}(\mathcal{A}^{-})$ & $w_{50}(\mathcal{A}^{-})$\\
        \midrule
        4B  & Pre-terminal & 2.319 & 2.454 & 36.18 & 36.90 \\
        30B & Pre-terminal & 1.851 & 1.735 & 19.50 & 18.80 \\
        \midrule
        \multicolumn{6}{@{}l}{\textbf{(b) Weight definition: raw versus length-normalized}}\\
        Model & Stage & $\rho_{\rm raw}$ & $\rho_{\rm eff}$ 
        & $w_{\rm raw}(\mathcal{A}^{-})$ & $w_{\rm eff}(\mathcal{A}^{-})$\\
        \midrule
        4B  & Healthy      & 1.023 & 0.654 & 25.11 & 19.98 \\
        4B  & Drift        & 1.504 & 0.796 & 28.92 & 20.83 \\
        4B  & Pre-terminal & 2.319 & 1.434 & 36.18 & 28.51 \\
        4B  & Terminal     & 1.936 & 1.297 & 28.38 & 25.30 \\
        30B & Healthy      & 1.337 & 0.963 & 23.98 & 20.30 \\
        30B & Drift        & 1.488 & 1.019 & 17.43 & 14.03 \\
        30B & Pre-terminal & 1.851 & 1.049 & 19.50 & 14.17 \\
        30B & Terminal     & 3.391 & 1.652 & 31.13 & 22.00 \\
        \bottomrule
    \end{tabularx}
\end{table}

Extending the pre-terminal window from 25 to 50 updates changes the
raw-weighted ratio from 2.319 to 2.454 on 4B and from 1.851 to 1.735
on 30B. Both windows retain the raw-weighted pre-terminal imbalance.
The intervals are 219--243 versus 194--243 for 4B and 519--543 versus
494--543 for 30B.

Length normalization changes the relative influence of long responses.
It reduces the pre-terminal ratio to 1.434 on 4B and 1.049 on 30B,
and changes whether the ratio exceeds one in some earlier stages.
Raw advantage-weighted token mass and length-normalized coefficient mass
therefore answer different questions. We retain the former for
Section~\ref{sec:directional_asymmetry} and report the latter explicitly
here. Both $\mathcal{A}^{-}$ and $\mathcal{A}^{+}$ are amplifying regions
under the diagonal sign criterion, so their ratio describes a directional
allocation within that part of the decomposition, not the balance between
all amplifying and contracting contributions.
\section{TRIAGE Specification and Derivations}
\label{app:gate_details}

This appendix specifies the segment gate, derives the response-level
reweighting in Section~\ref{sec:TRIAGE_gate}, and derives the repair
coefficient in Section~\ref{sec:TRIAGE_repair}. Throughout, advantages,
sampler log probabilities, masks, and gate weights are held fixed during
each learner update. The learner log probabilities used by repair retain
their gradients.

\subsection{Segment Construction and the Gate Operating Rule}
\label{app:gate_operating_rule}

Let $m_{i,t}\in\{0,1\}$ be the valid completion-token mask and
$n_i=\sum_t m_{i,t}$. Diagnostic segments contain consecutive response
positions and do not cross response boundaries. We retain the final shorter
segment and omit segments with no valid tokens. Masked positions contribute
neither to segment statistics nor to the loss. For a segment $S$, let
$n_S=\sum_{t\in S}m_{i,t}>0$. The two diagnostic statistics are those in
Eq.~\ref{eq:segment_statistics}.

For outcome-based GRPO, the advantage is constant within a response.
Consequently, a segment is eligible for gating exactly when $A_i<0$.
Two sources of risk determine its weight: a negative mean gap and a
concentration of deep negative-gap tokens. The mean-gap component uses
thresholds $\tau_{\mathrm{sev}}<\tau_{\mathrm{neg}}<0$ and weights
$0<w_{\mathrm{sev}}\leq w_{\mathrm{neg}}\leq1$:
\begin{equation}
 w_S^{\mathrm{mean}}=
 \begin{cases}
 w_{\mathrm{sev}}, & \bar\delta_S<\tau_{\mathrm{sev}},\\
 w_{\mathrm{neg}}, & \tau_{\mathrm{sev}}\leq\bar\delta_S<\tau_{\mathrm{neg}},\\
 1, & \bar\delta_S\geq\tau_{\mathrm{neg}}.
 \end{cases}
 \label{eq:app_gate_mean}
\end{equation}
The number of deep-tail tokens and the tail component are
\begin{equation}
 k_S^{\mathrm{tail}}=\sum_{t\in S}m_{i,t}
       \mathbf1[\delta_{i,t}<\tau_{\mathrm{tail}}]
       =n_S f_S^{\mathrm{tail}},\qquad
 w_S^{\mathrm{tail}}=1-(1-w_{\mathrm{sev}})
       \min\!\left(\frac{k_S^{\mathrm{tail}}}{K_{\mathrm{tail}}},1\right),
 \label{eq:app_gate_tail}
\end{equation}
where $K_{\mathrm{tail}}>0$ is the count at which tail attenuation reaches
$w_{\mathrm{sev}}$. Combining the two components gives
\begin{equation}
 w_S=
 \begin{cases}
 \min(w_S^{\mathrm{mean}},w_S^{\mathrm{tail}}), & A_i<0,\\
 1, & A_i\geq0.
 \end{cases}
 \label{eq:app_gate_rule}
\end{equation}
Thus $w_{\min}=w_{\mathrm{sev}}$. In the compact notation of
Eq.~\ref{eq:segment_risk_map}, $\mathcal G$ uses the segment's valid
count when converting $f_S^{\mathrm{tail}}$ to $k_S^{\mathrm{tail}}$.
This accounts for a short final segment or masked positions without
introducing a separate gate symbol.
Finally, Eq.~\ref{eq:direction_selective_gate} applies $w_S$ only to
negative-gap tokens in an eligible segment. The remaining tokens retain
raw weight one.

\subsection{From the Weighted Loss to the Effective Update}
\label{app:gate_reduction}

Write $H_i=\sum_t m_{i,t}w_{i,t}$ and
$\bar w_i=H_i/n_i$. Because the weights are positive on valid tokens,
$H_i>0$ for every nonempty response. Treating the gate as a constant during
backpropagation gives
\begin{equation}
 \nabla_\theta\mathcal L_{\mathrm{PG},i}^{\mathrm{TRIAGE}}
 =\frac1{H_i}\sum_t m_{i,t}w_{i,t}\nabla_\theta\ell_{i,t}^{\mathrm{base}}
 =\frac1{n_i}\sum_t m_{i,t}\frac{w_{i,t}}{\bar w_i}
                 \nabla_\theta\ell_{i,t}^{\mathrm{base}}.
 \label{eq:app_gate_gradient}
\end{equation}
This recovers $\alpha_{i,t}=w_{i,t}/\bar w_i$ in
Eq.~\ref{eq:effective_alpha}, including the original response and batch
reduction factors in $g_{i,t}^{\mathrm{base}}$. In particular,
\begin{equation}
 \frac1{n_i}\sum_t m_{i,t}\alpha_{i,t}=1,
 \qquad
 \frac{\alpha_{i,t}}{\alpha_{i,u}}=\frac{w_{i,t}}{w_{i,u}}
 \quad\text{for valid tokens }t,u.
 \label{eq:app_gate_normalization}
\end{equation}
The gate therefore changes relative allocation within a response.
An ungated token has $\alpha_{i,t}=1/\bar w_i\geq1$; an attenuated raw
weight need not yield $\alpha_{i,t}<1$. If every valid token receives the
same raw weight, all effective multipliers are one. The unit average of
$\alpha$ does not impose a conservation law on the parameter-gradient norm
or on the mismatch-energy contributions.

For comparison, a fixed-denominator variant uses
\begin{equation}
 \mathcal L_{\mathrm{PG},i}^{\mathrm{fixed}}
 =\frac1{n_i}\sum_t m_{i,t}w_{i,t}\ell_{i,t}^{\mathrm{base}},
 \qquad \alpha_{i,t}^{\mathrm{fixed}}=w_{i,t}.
 \label{eq:app_fixed_denominator}
\end{equation}
This variant changes the response's overall multiplier as well as its
within-response allocation.

\subsection{Repair Objective and Additive Coefficient}
\label{app:repair_details}

Repair uses non-overlapping segments $R$, with response index $i(R)$,
valid count $n_R=\sum_{t\in R}m_{i(R),t}>0$, and mean gap
\begin{equation}
 \bar\delta_R=\frac1{n_R}\sum_{t\in R}m_{i(R),t}\delta_{i(R),t},
 \qquad q_R=[\tau_{\mathrm{rep}}-\bar\delta_R]_+.
 \label{eq:app_repair_segment}
\end{equation}
Here $\delta_{i,t}=z_{i,t}(\theta)-\log\pi_{\mathrm s}
(y_{i,t}\mid x,y_{i,<t})$, using the same sampler notation as the main text.
Let $\mathcal S_{\mathrm{rep}}^{\mathrm{valid}}$ contain all nonempty repair
segments in the optimizer-update batch and
$N_{\mathrm{rep}}=|\mathcal S_{\mathrm{rep}}^{\mathrm{valid}}|$.
The denominator counts every such segment, including segments with
$q_R=0$ and segments from non-positive-advantage responses.
For $N_{\mathrm{rep}}>0$, define
\begin{equation}
 \mathcal L_{\mathrm{rep},k}
 =\frac{\lambda_k}{N_{\mathrm{rep}}}
   \sum_{R\in\mathcal S_{\mathrm{rep}}^{\mathrm{valid}}}
   \mathbf1[A_{i(R)}>0]\phi_\beta(q_R),
 \qquad
 \lambda_k=\lambda_{\mathrm{rep}}\mathbf1[k\geq k_{\mathrm{rep}}].
 \label{eq:app_repair_objective}
\end{equation}
When the valid segment set is empty, the repair term and its gradient are
zero. Counting all valid segments keeps the denominator independent of
which segments violate the repair boundary.

For a token $t\in R$ with $i=i(R)$ in an active segment, the sampler term is fixed and
\begin{align}
 \frac{\partial q_R}{\partial z_{i,t}}&=-\frac{m_{i,t}}{n_R},
 \label{eq:repair_q_grad}\\
 \phi_\beta'(q)&=\frac{q}{\sqrt{1+(q/\beta)^2}}.
 \label{eq:repair_phi_grad}
\end{align}
At $q_R=0$ the composed penalty has zero derivative. Let $R(i,t)$ be the
unique repair segment containing a valid token. Gradient descent on
Eq.~\ref{eq:app_repair_objective} contributes
\begin{align}
 \Delta\theta_{\mathrm{rep}}
 &=-\eta\nabla_\theta\mathcal L_{\mathrm{rep},k}
   =\eta\sum_{i,t}\psi_{i,t}\nabla_\theta z_{i,t},
 \label{eq:app_repair_update}\\
 \psi_{i,t}
 &=\frac{\lambda_k}{N_{\mathrm{rep}}}
   \frac{m_{i,t}}{n_{R(i,t)}}\phi_\beta'(q_{R(i,t)})
   \mathbf1[A_i>0,\ q_{R(i,t)}>0].
 \label{eq:repair_psi_appendix}
\end{align}
The coefficient is zero for tokens outside the valid repair set.
Equation~\ref{eq:repair_psi_appendix} recovers
Eq.~\ref{eq:repair_psi}, including the positive-advantage condition and
the activation schedule.

\paragraph{Bounded corrective coefficients.}
For $\beta>0$ and finite $q\geq0$, the pseudo-Huber penalty obeys
\begin{equation}
 \begin{aligned}
 \phi_\beta(q)&\sim\tfrac12q^2 &&(q/\beta\to0),\\
 \phi_\beta(q)&=\beta q-\beta^2+\mathcal O(\beta^3/q)
       &&(q/\beta\to\infty),\\
 0&\leq\phi_\beta'(q)<\beta.
 \end{aligned}
 \label{eq:app_repair_penalty_properties}
\end{equation}
Consequently, throughout training,
\begin{equation}
 0\leq\psi_{i,t}\leq
 \frac{\lambda_k\beta}{N_{\mathrm{rep}}n_{R(i,t)}}
 \label{eq:repair_bound}
\end{equation}
for valid tokens and nonempty repair sets. The upper inequality is strict
when $\lambda_k>0$. The non-strict form also covers the period before
repair is enabled. This bound controls the coefficient multiplying each
log-probability gradient; a bound on the parameter-gradient norm would
additionally depend on the norms of those gradients.

\paragraph{Repair on positive-advantage segments.}
When $\lambda_k>0$, an active positive-advantage segment with $q_R>0$
assigns the same positive coefficient to all valid tokens. Negative-gap tokens therefore have
$\delta_{i,t}\psi_{i,t}<0$ in the diagonal decomposition of
Section~\ref{sec:mismatch_dynamics}. Positive-gap tokens in the same segment
also receive this coefficient. Thus the objective corrects negative
segment-level displacement without imposing contraction on every token.
\subsection{Implementation and Configuration}
\label{app:triage_implementation}

Gate diagnostics and weights are detached. The positive-advantage repair
branch retains live learner gaps for valid tokens in a segment with
$A_i>0$. For the remaining tokens, detached copies preserve the segment
mean used by the diagnostic computation. The loss numerator then includes
only selected positive-advantage segments, while the denominator counts
all valid repair segments. Both the resulting repair value and its learner
gradient therefore match Eq.~\ref{eq:app_repair_objective}.

The activation boundary is inclusive: repair is enabled when the persisted
rollout-update index reaches $k_{\mathrm{rep}}$. With one learner update
per rollout, this index supplies the schedule in
Eq.~\ref{eq:app_repair_objective}. The schedule changes the repair
coefficient; gate diagnostics, segment counts, and the policy objective
are unchanged before activation.

To implement the response-mean policy objective in
Eq.~\ref{eq:TRIAGE_pg_loss}, first form each response's weighted mean
and then apply the outer response average. The repair denominator is
computed over the full optimizer-update batch. If the loss interface
subsequently divides accumulated numerators by one denominator for the complete update,
scale each local selected repair sum by that denominator divided by
$N_{\mathrm{rep}}$ before the outer division. The result is the same global
segment mean regardless of the data-parallel or microbatch partition.

\begin{table}[t]
    \centering
    \small
    \setlength{\tabcolsep}{5pt}
    \caption{
    Gate and repair configuration.
    The diagnostic segment length matches the 64-token length used
    in characterization; $\tau_{\mathrm{ana}}$ remains a separate
    characterization threshold.
    Repair starts inclusively at the persisted, zero-based
    rollout-update index $k_{\mathrm{rep}}$.
    }
    \label{tab:triage_configuration}
    \begin{tabular}{@{}llr@{}}
        \toprule
        Parameter & Meaning & Value \\
        \midrule
        $W_{\mathrm{gate}}$
        & Diagnostic segment length & $64$ \\
        $\tau_{\mathrm{neg}}$
        & Negative mean-gap threshold & $-0.5$ \\
        $\tau_{\mathrm{sev}}$
        & Severe mean-gap threshold & $-1.5$ \\
        $\tau_{\mathrm{tail}}$
        & Deep-tail token threshold & $-6.0$ \\
        $w_{\mathrm{neg}}$
        & Negative mean-gap weight & $0.3$ \\
        $w_{\mathrm{sev}}$
        & Severe/minimum raw weight & $0.1$ \\
        $K_{\mathrm{tail}}$
        & Tail-count saturation & $2$ \\
        \midrule
        $W_{\mathrm{rep}}$
        & Repair segment length & $128$ \\
        $\tau_{\mathrm{rep}}$
        & Mean-gap repair boundary & $-0.03$ \\
        $\beta$
        & Pseudo-Huber scale & $0.022$ \\
        $\lambda_{\mathrm{rep}}$
        & Repair coefficient after activation & $0.015$ \\
        $k_{\mathrm{rep}}$
        & Inclusive repair start index & $0$ \\
        \bottomrule
    \end{tabular}
\end{table}

\paragraph{One learner update.}
The following procedure implements the objective in
Eq.~\ref{eq:total_TRIAGE_loss}:
\begin{enumerate}[leftmargin=*,itemsep=2pt,topsep=2pt]
 \item Obtain learner log probabilities on the sampled prefixes and subtract
 the fixed generation-time sampler log probabilities. Retain a live gap
 for repair and a detached copy for gate diagnostics.
 \item Partition each response into diagnostic segments, compute the masked
 statistics, and evaluate Eqs.~\ref{eq:app_gate_mean}--\ref{eq:app_gate_rule}.
 Apply the resulting weights only to the negative-gap tokens of
 negative-advantage responses.
 \item Compute each response's weighted policy mean and average these
 response losses using the baseline's batch reduction.
 \item Partition responses into repair segments and count every nonempty
 segment across the optimizer-update batch. Set $\lambda_k$ from the fixed
 schedule and evaluate the positive-advantage repair objective.
 \item Add the two losses and take the learner optimizer step. The effective
 current-gradient coefficient is
 $g_{i,t}=\alpha_{i,t}g_{i,t}^{\mathrm{base}}+\psi_{i,t}$; optimizer-history
 effects are represented by the residual in Appendix~\ref{app:mismatch_derivation}.
\end{enumerate}

\section{Training and Evaluation Protocols}
\label{app:training_details}

\subsection{Main Training Comparisons}\label{app:main_training}
The main comparisons in Section~\ref{sec:experiments} evaluate training
stability and downstream reasoning performance with Qwen3-30B-A3B-Base
as the primary model and Qwen3-4B-Base for replication and ablation.
We use the DAPO dataset for training and GRPO for policy optimization.
The characterization runs in Section~\ref{sec:understanding} instead
provide the fixed trajectories used for directional and localization
analysis. Their run-specific horizons, collapse criteria, and mismatch
estimands are specified in Appendix~\ref{app:analysis_setup:runs} and
Appendix~\ref{app:analysis_setup:estimands}.

The main training configurations are BF16, native NVFP4, NVFP4 with
token-level truncated importance sampling (TIS), and NVFP4 with TRIAGE.
The TIS truncation threshold is $C=2.0$. TRIAGE applies the gating and
repair operations in Section~\ref{sec:method} to the importance-corrected
objective. The native NVFP4 configurations use W4A4 forward execution on
the sampler and learner, with the activation-scaling recipe described in
Section~\ref{sec:experimental_setup}. The maximum response length in the
main training setup is $20{,}480$ tokens.

\subsection{Downstream Evaluation}
\label{app:evaluation_details}

We evaluate mathematical reasoning on AIME 2024, AIME 2025, MATH500,
AMC23, and GSM8K. Table~\ref{tab:benchmark_eval} identifies the training
step of each evaluated checkpoint. Its $\dagger$ marker denotes a run
that collapses before its planned budget; the reported result then uses
the last saved checkpoint preceding collapse. The Qwen3-4B-Base TRIAGE
results include the intermediate step-300 checkpoint and the final
step-600 checkpoint.

To summarize the five benchmark scores, we use the unweighted mean
\begin{equation}
    \operatorname{Avg}
    =\frac{1}{5}\sum_{b=1}^{5}\operatorname{Score}_b,
    \label{app:eq:benchmark_average}
\end{equation}
where each $\operatorname{Score}_b$ is expressed as a percentage. This
gives each benchmark equal weight, independently of its number of
problems. Training reward, rollout pass rate, and the mean absolute
learner--sampler log-probability gap are tracked separately as training
diagnostics in Figure~\ref{fig:main_curves}.

\section{Efficiency and Profiling}
\label{app:efficiency}

\subsection{Natural-Generation RL Measurements}
\label{app:natural_rl_efficiency}

The rollout-throughput measurements in Table~\ref{tab:efficiency} use
Qwen3-30B-A3B-Base on one node with eight NVIDIA Blackwell B300 GPUs. Slime
coordinates colocated Megatron-LM learner and SGLang sampler execution.
The learner parallel configuration is tensor, pipeline, and context
parallel size 1, expert parallel size 8, and expert tensor parallel
size 1. Each prompt has 16 sampled responses. Global batch sizes 256
and 512 therefore correspond to 16 and 32 prompts, respectively.

Each run starts from the base checkpoint and trains on
\textsc{DAPO-Math-17K} with seed 42 and the same data order. Responses
terminate naturally at EOS or at the $20{,}480$-token limit. A run
contains 12 rollout iterations: iterations 0--1 are warm-up and
iterations 2--11 supply the ten measured observations. Initialization,
evaluation, and checkpoint saving are excluded.

For measured iteration $j$, let $M_j$ be the number of generated
completion tokens and $T_j^{\mathrm{rollout}}$ the rollout duration.
The reported throughput is the arithmetic mean of per-iteration
throughputs,
\begin{equation}
    \overline{\mathrm{TPS}}
    =\frac{1}{10}\sum_{j=2}^{11}
      \frac{M_j}{T_j^{\mathrm{rollout}}}.
    \label{app:eq:rollout_throughput}
\end{equation}
This counts the actual output tokens produced by each precision
configuration. Iteration wall time includes rollout, learner updates,
weight conversion and synchronization, and orchestration. Learner
timing includes log-probability recomputation and the backward and
optimizer update. In colocated execution, overlapping wait intervals
are not added again to the wall-time total.

\subsection{Paired Learner-Overhead Measurement}\label{app:learner_overhead}

The incremental learner cost of TRIAGE is measured separately on
eight B300 GPUs with five paired repetitions of the same frozen
batch. Gating and repair are enabled for one member of each pair.
The observed median learner-update overhead is $0.84\%$, with a
median absolute increase of $0.225$\,s. This comparison holds the
training workload fixed when measuring the additional objective
computation; natural-generation iteration times also reflect changes
in output lengths.

\subsection{Fixed-output sampler evaluation}\label{app:fixed_output_sampler}
\subsubsection{Independent Fixed-Output Configuration Selection}
\label{app:fixed_output_efficiency}

We additionally measure standalone sampler throughput with a fixed
output workload. For Qwen3-30B-A3B, each batch contains 512 responses
of 4,096 output tokens with a client concurrency limit of 512 on one
B300 GPU. Throughput divides the total output tokens by the complete
client batch time, including queuing and final request completion.
The KV cache uses BF16. This sampler experiment uses PyTorch
2.9.1+cu130, SGLang 0.5.10, FlashInfer 0.6.12, and
\texttt{sglang-kernel} 0.4.1+cu130.

\begin{table}[htbp]
    \centering
    \caption{\textbf{Standalone Qwen3-30B-A3B throughput after configuration selection.} Ranges are min--max over five repetitions. Speedup is the NVFP4/BF16 median-throughput ratio.}
    \label{tab:fixed}
    \small
    \begin{tabular}{@{}llrrrr@{}}
    \toprule
    Precision & Mem. frac. & Median & Range & Speedup\\
     & & & \multicolumn{2}{c}{(output tokens/s)} & \\
    \midrule
    BF16  & 0.95 & 15,470.01 & 15,447.56--15,509.83 & 1.000$\times$\\
    NVFP4  & 0.90 & 18,175.67 & 18,126.37--18,195.69 & 1.175$\times$\\
    \bottomrule
    \end{tabular}
\end{table}

Both precisions receive a 12-candidate configuration search. The two leading candidates are remeasured, and the selected configuration for each precision is measured five times on the same physical GPU. Both selected configurations use the FlashInfer TRT-LLM MoE backend. The selected static memory fractions are 0.95 for BF16 and 0.90 for NVFP4. Table~\ref{tab:fixed} reports the resulting medians and observed ranges. Under these independently selected configurations, NVFP4 throughput is 1.175$\times$ the BF16 throughput.

\subsubsection{Concurrency and output length}
\begin{figure}[htbp]
    \centering
    \includegraphics[width=1.0\linewidth]{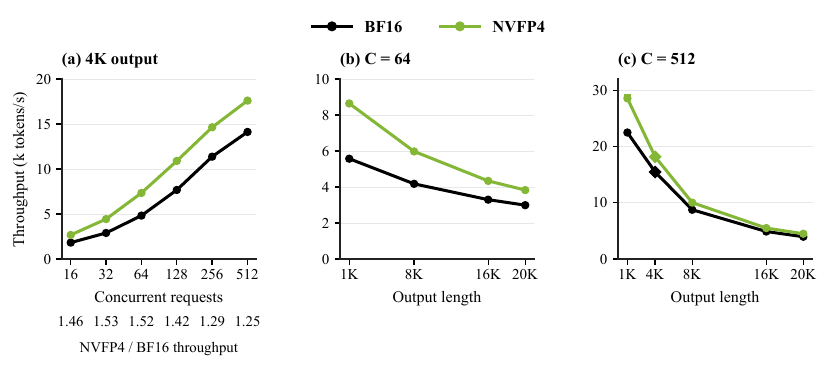}
    \caption{\textbf{Fixed-output serving sweeps on Qwen3-30B-A3B.} Each batch contains 512 responses on one B300 GPU. (a) Concurrency at 4K output tokens. The row below gives NVFP4/BF16 throughput ratios. (b,c) Length at concurrency 64 and 512. K denotes 1,024 tokens. Points show medians and whiskers min--max over three runs, except the separately confirmed C512/4K point (diamond), which uses five. The common-configuration C64/4K point is omitted from the selected-configuration length curve.}
    \label{fig:scans}
\end{figure}

Figure~\ref{fig:scans} varies the serving operating point with fixed output workloads. The concurrency sweep uses a common starting configuration with memory fraction 0.8. Length sweeps retain the precision-specific selected configurations in Table~\ref{tab:fixed}.
Throughput increases with concurrency over the measured 4K sweep. At fixed concurrency, it decreases as output length grows, and the relative gain varies with the operating point. These measurements complement Section 5.3 by holding the output workload fixed.

\subsection{Controlled Long-Tail Workload}\label{app:longtail_workload}
We isolate response-length heterogeneity with a fixed request-to-length mapping on eight independent single-GPU Qwen3-30B-A3B sampler replicas. At batch size 256, the mixed workload contains 224 responses of 2,048 tokens, 24 of 8,192 tokens, and eight of 20,480 tokens. Batch size 512 doubles each count. A uniform control assigns 3,200 tokens per response, preserving the total output-token count. The mapping uses seed 20260912, with identical requests and target lengths for both precisions.

For mixed lengths, median batch time decreases from 156.13 to 120.83~s at batch size 256 and from 165.26 to 129.28~s at batch size 512. The time ratios are 1.292$\times$ and 1.278$\times$. The equal-token uniform controls yield larger ratios of 1.538$\times$ and 1.513$\times$.

NVFP4 therefore reduces absolute waiting time in these long-tail batches by approximately 35--36~s, while the relative gain is smaller than for uniform outputs. The completion curves expose the late portion of the batch that average throughput alone does not describe. These controlled sampler workloads complement Section 5.3 by fixing the output-length assignment.
\begin{figure}[tbp]
    \centering
    \includegraphics[width=1.0\linewidth]{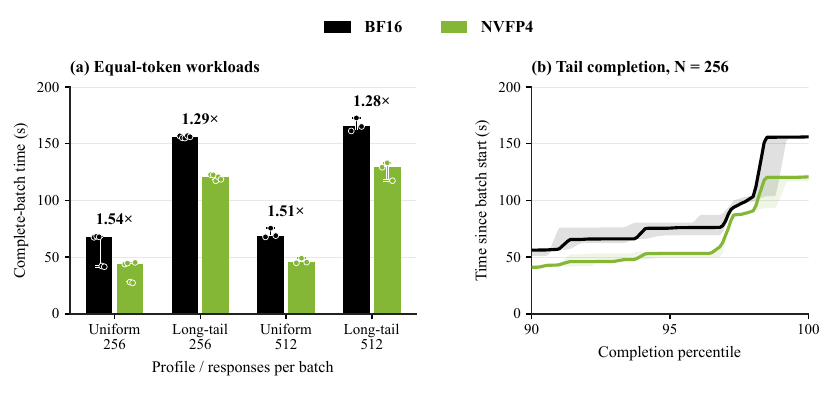}
    \caption{\textbf{Controlled long-tail and equal-token uniform workloads.} (a) Complete-batch wall time. Bars show medians, points individual runs, and whiskers min--max. Labels are BF16/NVFP4 median-time ratios. (b) Completion-time quantiles for the final 10\% of requests in the 256-response mixed workload, with pointwise median and min--max across five repetitions. Batch size 256 uses five repeats per condition and batch size 512 uses three. Timing includes request handling, returned log-probabilities, and final draining, excluding initialization and warm-up.}
    \label{fig:longtail}
\end{figure}

\end{document}